%% file: segmentation_RF.tex
\documentclass[journal]{IEEEtran}
\input{commands}

\usepackage{xcolor}
\usepackage{colortbl}
\usepackage{algorithm}
\usepackage{algorithmic}
\usepackage{bbm}
\usepackage{mathtools}
\newcommand{\etal}{{\it{et al}.} }

\begin{document}
\title{Random Forest with Learned Representations \\for Semantic Segmentation}

\author{Byeongkeun~Kang
        and~Truong~Q.~Nguyen,~\IEEEmembership{Fellow,~IEEE}
\thanks{B. Kang and T. Q. Nguyen are with the Department of Electrical and Computer Engineering, University of California, San Diego, CA 92093 USA (e-mail: bkkang@ucsd.edu and tqn001@eng.ucsd.edu).}
\thanks{This work is supported in part by NSF grant IIS-1522125.}
}
\maketitle

\begin{abstract}
In this work, we present a random forest framework that learns the weights, shapes, and sparsities of feature representations for real-time semantic segmentation. Typical filters (kernels) have predetermined shapes and sparsities and learn only weights. A few feature extraction methods fix weights and learn only shapes and sparsities. These predetermined constraints restrict learning and extracting optimal features. To overcome this limitation, we propose an unconstrained representation that is able to extract optimal features by learning weights, shapes, and sparsities. We, then, present the random forest framework that learns the flexible filters using an iterative optimization algorithm and segments input images using the learned representations. We demonstrate the effectiveness of the proposed method using a hand segmentation dataset for hand-object interaction and using two semantic segmentation datasets. The results show that the proposed method achieves real-time semantic segmentation using limited computational and memory resources.
\end{abstract}

\

\IEEEpeerreviewmaketitle

\input{1_introduction}

\input{2_related_works}
\input{3_proposed_method}

\input{3_3_random_forest}
\input{3_4_filtering}
\input{4_experiments_results}

\section{Conclusion} \label{sec:conclusion}
In this paper, we present the random forest framework that employs the novel unconstrained representation to achieve real-time semantic segmentation. The unconstrained representation is proposed to learn optimal features to achieve higher accuracy. The random forest framework is selected to achieve high efficiency for real-time processing. The results verify that the proposed method achieves higher accuracy comparing to previous random forest frameworks and processes using much smaller computational and memory resources comparing to deep learning-based methods.

\bibliographystyle{IEEEtran}
\bibliography{segmentation_RF}

\vfill
\end{document}

%% file: commands.tex
\usepackage{subfigure}
\usepackage{cite}
\usepackage[pdftex]{graphicx}
\graphicspath{{../pdf/}{../jpeg/}}
\DeclareGraphicsExtensions{.pdf,.jpeg,.png}
\usepackage{amsmath}
\usepackage{algorithmic}
\usepackage{array}
\ifCLASSOPTIONcompsoc
  \usepackage[caption=false,font=normalsize,labelfont=sf,textfont=sf]{subfig}
\else
  \usepackage[caption=false,font=footnotesize]{subfig}
\fi
\usepackage{url}

\usepackage{amssymb}
\usepackage{multirow}
\usepackage{tabu}
\usepackage{pbox}

{\begin{list}               
    {$\bullet$ \hfill}{
        \setlength{\leftmargin}{\parindent}
        \setlength{\parsep}{0.04\baselineskip}
        \setlength{\itemsep}{0.5\parsep}
        \setlength{\labelwidth}{\leftmargin}
        \setlength{\labelsep}{0em}}
    }
{\end{list}}

\providecommand{\eref}[1]{\eqref{#1}}  
\providecommand{\cref}[1]{Chapter~\ref{#1}}
\providecommand{\sref}[1]{Section~\ref{#1}}
\providecommand{\fref}[1]{Fig.~\ref{#1}}
\providecommand{\tref}[1]{Table~\ref{#1}}

\providecommand{\R}{\ensuremath{\mathbb{R}}}

\providecommand{\Z}{\ensuremath{\mathbb{Z}}}

\providecommand{\abs}[1]{\lvert#1\rvert}
\providecommand{\norm}[1]{\lVert#1\rVert}

\renewcommand{\vec}[1]{\ensuremath{\boldsymbol{#1}}}
\providecommand{\mat}[1]{\ensuremath{\boldsymbol{#1}}}

\providecommand{\calN}{\mathcal{N}}

\providecommand{\calT}{\mathcal{T}}

\providecommand{\calX}{\mathcal{X}}

\providecommand{\mD}{\mat{D}}

\providecommand{\mX}{\mat{X}}
\providecommand{\mY}{\mat{Y}}

\providecommand{\vp}{\vec{p}}
\providecommand{\vq}{\vec{q}}

\providecommand{\vu}{\vec{u}}

\providecommand{\vx}{\vec{x}}

\providecommand{\ptilde}{\widetilde{p}}

\providecommand{\ti}{\widetilde{i}}
\providecommand{\ttt}{\widetilde{t}}

\newcommand{\argmin}[1]{\mathop{\underset{#1}{\mbox{argmin}}}}


%% file: 1_introduction.tex
\section{Introduction}

\IEEEPARstart{A}{ccurate} and efficient semantic segmentation is a fundamental task in a variety of computer vision applications including autonomous driving, human-machine interaction, and robot navigation. Reducing computational complexity and memory usage is important to minimize response time and power consumption for portable devices such as robots and virtual/augmented devices. Moreover, it is beneficial for vehicles and robots to navigate in actively changing environments and for human-machine interaction devices to communicate without delay. However, it is challenging to achieve accurate and efficient semantic segmentation because every pixel needs to be classified using limited computational resources. 

To achieve accurate and efficient pixel-wise classification, Shotton \etal presented a random forest-based method and applied it for semantic segmentation and body pose estimation in~\cite{Shotton, shottoncvpr, shottonpami}. This work has been broadly employed in many related applications~\cite{Ren, sharp, tompson, kangglobalsip} and in Microsoft Kinect~\cite{kinect}. To improve accuracy in semantic segmentation, convolutional neural network-based methods have been proposed in~\cite{longcvpr, YuKoltun2016, Zheng, Chen, Chen16, kangtmm}. Part of the reason that deep learning-based methods outperform other methods is its ability to learn fine representations along with the hierarchical structure and non-linear activations. The methods using random forest and using deep neural networks outperform most of the other methods in the task of semantic segmentation in terms of efficiency and accuracy. Between two approaches, random forest-based methods have an advantage in computational complexity and memory usage while deep learning-based methods achieve higher accuracy. Therefore, we propose a random forest framework that employs unconstrained representations, learns optimal features by using an optimization algorithm, and inferences in real-time using limited computational and memory resources.

The proposed method can be applied to any input (e.g. color image, depth map, point cloud) and for any pixel-wise classification task. We validate the proposed method in the task of semantic segmentation and hand segmentation for hand-object interaction. Semantic segmentation is needed in many applications such as autonomous driving and robot navigation~\cite{cityscapes, Geiger2012CVPR, Geiger2013IJRR, BrostowSFC:ECCV08, BrostowFC:PRL2008}. We apply the proposed method to a road scene dataset~\cite{cityscapes} and an indoor scene dataset~\cite{nyudv2}, and use color images along with disparity/depth maps as input. Hand segmentation is also a fundamental task in human-machine interaction that is demanded in virtual reality (VR), augmented reality (AR), robotics, and user interfaces in an automobile~\cite{tzionas, cai, Ren, wang, wangtog, sharp, romeroivc, romeroicra, tompson, qian, oikoiccv, kangisvc, kangacpr, 3dhandpose, li, deephand, parsingHand, handTracking}. We apply the method to a depth map-based hand segmentation dataset for hand-object interaction~\cite{kangtmm}. In experiments, we show that the proposed method can be applied to real-time semantic segmentation task using limited computational and memory resources. 

\begin{figure}[!t]
\centering
   \includegraphics[width=0.48\textwidth]{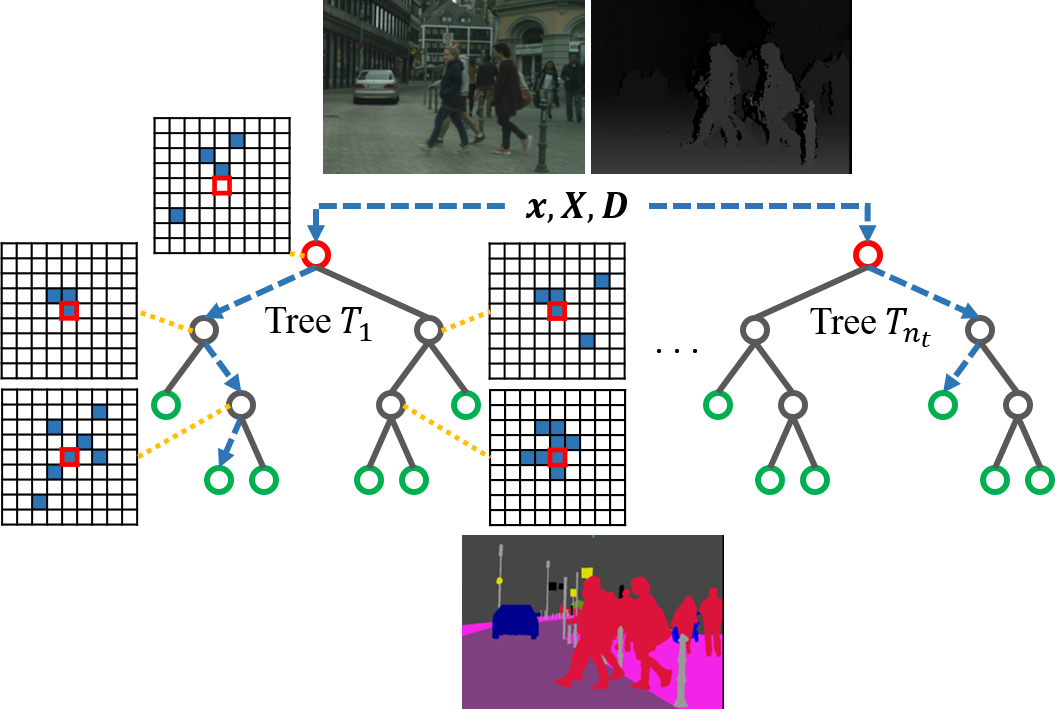}
   \caption{Illustration of the random forest with learned representations for semantic segmentation. The top images show input images (a color image and a depth map). The bottom image shows an expected output (class labels for each pixel). The rectangles with grid show examples of learned representations with various shapes, sparsities, and weights. Red bounding boxes present input data points, and blue color squares denote the offset points that are used to compute features.}
\label{fig:Intro}
\end{figure}

In this paper, we propose the novel unconstrained representation that is able to learn weights, shapes, and sparsities in~\sref{subsec:feature}. Then, we propose a random forest framework that learns the weights, shapes, and sparsities of the representation and inferences input data using the learned representation in~\sref{subsec:RF}. The section explains selecting training data using bootstrap aggregation and boosting, learning splitting functions using particle swarm optimization, and inferencing input images. In~\sref{sec:result}, we demonstrate the effectiveness of the proposed method on three tasks: road scene semantic segmentation, indoor scene semantic segmentation, and hand segmentation for hand-object interaction. We use publicly available Cityscapes dataset~\cite{cityscapes}, NYUDv2 dataset~\cite{nyudv2}, and HOI dataset~\cite{kangtmm}. 

In summary, the contributions of our work are as follows:
\begin{itemize}
  \item We propose the unconstrained representation that is able to represent any weights, shapes, and sparsities.
  \item We develop the random forest framework that learns the weights, shapes, and sparsities of the representation by using particle swarm optimization.
  \item We verify the effectiveness and efficiency of the proposed method on semantic segmentation and hand segmentation tasks.
\end{itemize}

%% file: 2_related_works.tex
\section{Related Works} \label{subsec:relatedWork}
\begin{figure}[!t]
\centering
   \includegraphics[width=0.48\textwidth]{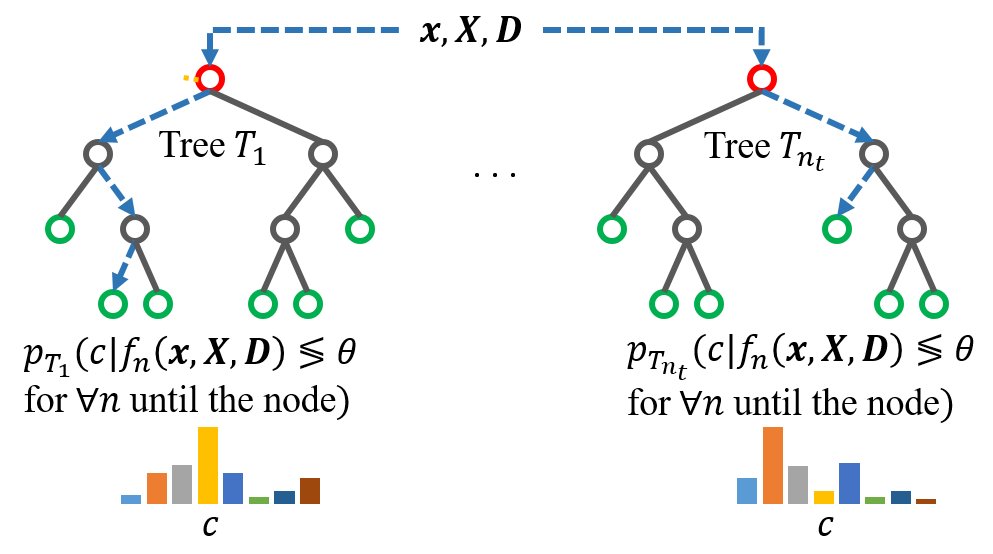}
   \caption{Random forest. Red, black, and green circles denote root nodes, split nodes, and leaf nodes, respectively. At leaf nodes, the forest estimates and uses the conditional probability of being each class given the specific leaf node.}
\label{fig:RF}
\end{figure}

\textbf{Per-pixel classification using random forest.} 
Random forest is an ensemble learning method and consists of a set of decision trees~\cite{Ho1995, Breiman2001, RFBook} as shown in Figs.~\ref{fig:Intro} and~\ref{fig:RF}. It is robust to noisy and variant data because of the combination of multiple trees with varying features and splitting criteria. Also, it is computationally less complex than typical neural networks. Part of the reason is that an input data is processed only log-scale portion of each tree based on the conditions in the ancestral nodes (see blue dotted lines in Figs.~\ref{fig:Intro} and~\ref{fig:RF}).

Shotton \etal presented semantic texton forest for image categorization and semantic segmentation~\cite{Shotton}. In order to avoid expensive computations of local descriptors (e.g. HOG~\cite{hog}, SIFT~\cite{shift}) or filter-bank responses, they employed splitting functions using the value of a single pixel, the sum, the difference, and the absolute difference of a pair of pixels. The method was extremely fast comparing to k-means clustering or nearest-neighbor assignment using feature descriptor. Schroff \etal investigated using not only local features but also global and context-rich features in the random forest for semantic segmentation~\cite{Schroff}. They showed that combining multiple features improves accuracy, and further demonstrated that relaxing constraints on features leads to higher classification accuracy. Shotton \etal extended the random forest~\cite{Shotton, RFKeypoint} to real-time body pose estimation by classifying each pixel to body parts~\cite{shottoncvpr, shottonpami}. They employed the depth difference between a pair of pixels as a feature. The feature fulfilled depth invariance property by calculating the location of the pixels considering depth information. This work was extended to hand segmentation by Tompson \etal~\cite{tompson}. Sharp \etal employed the memory-efficient random forest method~\cite{shottonJungle} to predict initial hand pose in hand tracking~\cite{sharp}. Kang \etal proposed the two-stage random forest method consisting of detection and segmentation for hand segmentation in hand-object interaction~\cite{kangglobalsip}.

The earlier works used hand-crafted features such as local descriptors (e.g. HOG, SIFT) or filter-bank responses~\cite{RFVoting}. Then, relatively recent works used pixel value difference as a feature that learns offsets while using fixed weights (+1 and -1)~\cite{RFKeypoint, shottoncvpr, shottonpami, tompson, kangglobalsip}. As investigated on the benefits of relaxing constraints of features~\cite{Schroff}, we further relax constraints of feature representations and propose the unconstrained representation. We enable learning optimal representation by learning weights, shapes, and sparsities of the proposed representation. Then, the learned feature is employed in the random forest framework for efficiency to achieve real-time inferencing.  

\textbf{Per-pixel classification using deep learning.} 
Deep learning-based methods achieved state-of-the-art accuracy recently in per-pixel classification.  Long \etal proposed fully convolutional neural networks~(FCN) for semantic segmentation by converting fully connected layers to convolutional layers in the neural networks for image classification~\cite{longcvpr, longpami}. Consequently, it takes an input of arbitrary size and produces an output of the corresponding size with pixel-wise prediction. Additional efforts have been made to achieve higher accuracy in~\cite{Zheng, Chen, Chen16, YuKoltun2016, kangtmm, nonlocal}. Zheng \etal proposed the convolutional neural networks that combine the strength of convolutional neural networks and conditional random field (CRF)-based probabilistic graphical modeling. They formulated CRF as recurrent neural networks and attached the recurrent neural networks following FCN~\cite{Zheng}. Chen \etal improved semantic segmentation using convolution with upsampled filters, atrous spatial pyramid pooling, and fully connected CRF~\cite{Chen, Chen16}. Yu \etal proposed an additional context module to aggregate multiscale information without losing resolution~\cite{YuKoltun2016}. Kang \etal proposed the depth-adaptive deep neural network that compensates depth variation to achieve depth-invariance property for semantic segmentation~\cite{kangtmm}. Wang \etal proposed the non-local operation that captures long-range dependencies to improve the accuracy of object detection/segmentation as well as video classification~\cite{nonlocal}.

Several other neural networks were proposed considering computational complexity~\cite{segnet, enet}. Badrinarayanan \etal proposed an efficient convolutional neural network in terms of memory and computational time during inference~\cite{segnet}. The network consists of an encoder network and a decoder network. The encoder network records pooling indices in max-pooling steps and the corresponding decoder network uses the recorded indices to perform nonlinear upsampling. Chaurasia \etal presented an efficient neural network to achieve real-time semantic segmentation~\cite{enet}. They analyzed various network architectures including early downsampling, nonlinear operations, and factoring filtering. 

While neural network-based methods achieve state-of-the-arts accuracy by learning optimal weights and biases in multiple layers with nonlinear activation functions and by extracting meaningful information from input data, they demand high computational and memory resources. Hence, we aim to present a random forest-based framework that is able to process real-time semantic segmentation using only limited computational and memory resources (e.g. low-end GPU or embedded system). Moreover, while typical neural networks use a hand-picked shape and sparsity~\cite{Zheng, Chen, Chen16, YuKoltun2016}, we learn shapes and sparsities along with weights by using an optimization algorithm.



\textbf{Learning representation.}
In random forest, learning two offset points for a feature representation has been one of the most popular representation learning methods~\cite{RFKeypoint, shottoncvpr, shottonpami, tompson, kangglobalsip}. While it can learn any offset vectors to represent various shapes and sparsities, it has constraints of fixed weights (+1 or -1) and of using only two offset points.

Typical neural networks learn representations using dense filters with various filter-sizes~\cite{lenet, AlexNet, vgg}. Recently, dilated convolution (also known as atrous convolution) was introduced to learn sparse representations~\cite{Chen16}. The sparse convolution was also applied in depth-adaptive convolution to learn depth-invariant representations~\cite{kangtmm}. While these representations have square shapes, active convolution and deformable convolution were presented to learn representations with various shapes in~\cite{activeConv, deformConv}. Both methods learn shapes of convolution filters using a training dataset. Active convolution defines learnable position parameters to represent various forms of receptive fields~\cite{activeConv}. Deformable convolution uses the offset field similar to the position parameters~\cite{deformConv}. It computes the offset field using the input feature map at each spatial location.

In this work, we aim to learn weights, shapes, and sparsities of descriptors by proposing an unconstrained representation and by applying particle swarm optimization. Comparing to learning two offset points in random forest, we also learn weights and the number of offset points. Concerning the recent convolution layers, we further learn the number of offset points. Moreover, the proposed representation has depth-invariance property as offset points are computed with the consideration of the distance from a camera.

%% file: 3_proposed_method.tex
\section{Proposed method}
In this section, we introduce an unconstrained representation that is able to represent any weights, shapes, and sparsities in~\sref{subsec:feature}. The unconstrained filter is proposed to learn optimal filters for a given task and dataset during training. We employ the proposed representation in the framework of random forest in~\sref{subsec:RF}. We describe details about training procedure including learning optimal representations and selecting training data points in~\sref{subsec:training} and about inference method in~\sref{subsec:inference}.

\subsection{Notation} \label{sec:notation}
Let $\mX \in \R^{p \times q}$ and $\mY \in \R^{p \times q \times n_c}$ be the matrices denoting an input image and an output probability map of a random forest where $p$, $q$, and $n_c$ represent the height, the width, and the number of classes, respectively. At a location $\vx \in \R^{2}$ on the image $\mX$, the intensity is represented as $\mX_{\vx}$. Also, let $\calT$, $T_i$, and $n_t$ indicate the random forest, the $i$-th decision tree, and the number of trees in the forest.
\begin{equation}
\calT = \{T_i | i \leq n_t \ \text{and } i \in \Z^{+} \}.
\label{eq:forest}
\end{equation}

\input{fig_RF_feature}

\subsection{Unconstrained Representation} \label{subsec:feature}
Learning optimal representation is essential to achieve higher accuracy and to avoid unnecessary use of computation and memory. Hence, we design an unconstrained representation that is able to learn optimal weights, shapes, and sparsities. In the proposed framework, the unconstrained representation's weights, shapes, and sparsities are learned at each node to split data points of different classes to separate child nodes as shown in~\fref{fig:Intro}. Moreover, the representation extracts depth and shift invariant features by compensating the associated changes as described in~\fref{fig:Feature}. Hence, the proposed novel representation improves accuracy and reduces computational complexity and memory usage.

\input{fig_feature_shape}

Given an input data (spatial location) $\vx$ and the corresponding input image and depth map ($\mX$, $\mD$), the feature $f(\cdot)$ is defined as follows:
\begin{equation}
f(\vx, \mX, \mD) = \sum_{i=1}^{n_f} w_i \mX_{\vx+{\vu_i}/{\mD_{\vx}}, h}
\label{eq:feature1}
\end{equation}
where $n_f$ is the number of data points used to compute each feature; $w$ and $\vu \in \R^{2}$ are a weight and an offset parameter vector, respectively; $h$ is the channel index of $\mX$. 

In details, the offset data point $\vx+{\vu}/{\mD_{\vx}}$ (the other end-point from a red circle $\vx$ in~\fref{fig:Feature}) is determined based on $\vx$, $\vu$, and $\mD_{\vx}$. The weight $w$ controls the influence of the information at the offset data point $\mX_{\vx+{\vu}/{\mD_{\vx}}}$. The feature response is computed by the weighted summation of the offset data points. All the parameters including the number $n_f$ of data points, the weights $w$, the offsets $\vu$, and the channel $h$ are learned in the training procedure in~\sref{subsec:training}. Each learned representation consists of $4 \times n_f + 1$ parameters where the factor 4 includes a weight, two offset parameters, and a channel index and the addition of 1 is for the number of data points. The proposed representation is described in Figs.~\ref{fig:Feature} and~\ref{fig:FeatureShape}.

If any offset data point $\vx+{\vu}/{\mD_{\vx}}$ is beyond the boundary of the image, the intensity $\mX_{\vx+{\vu}/{\mD_{\vx}}}$ is replaced by a constant (the maximum intensity of the input image $\mX$ if $\mX$ is a depth map; 0 otherwise).
\begin{equation}
\begin{split}
\mX_{\vx+{\vu}/{\mD_{\vx}}} = \begin{cases}
                                \mX_{\vx+{\vu}/{\mD_{\vx}}} &\text{ if }  \vx+{\vu}/{\mD_{\vx}} \in \mathbb{Z}^+ \text{ and } \\
                                                   &\text{ \quad         }         \vx+{\vu}/{\mD_{\vx}} \leq (p,q), \\
                                 \max(\mX) &\text{ else if } \mX \text{ is a depth map},  \\
                                 0 &\text{ otherwise.} \\
                                \end{cases}
\end{split}
\end{equation}

Obviously, when an input image $\mX$ is a depth map $\mD$, the feature can be expressed as follows: 
\begin{equation}
f(\vx, \mX) = \sum_{i=1}^{n_f} w_i \mX_{\vx+{\vu_i}/{\mX_{\vx}}}.
\end{equation}

Additionally, if depth information is not available, the requirement of depth data can be relaxed by sacrificing the property of depth invariance. The relaxed feature is defined as follows:
\begin{equation}
f(\vx, \mX) = \sum_{i=1}^{n_f} w_i \mX_{\vx+\vu_i}.
\label{eq:feature2}
\end{equation}

%% file: fig_RF_feature.tex
\begin{figure}[!t] \begin{center}
\begin{minipage}{0.49\linewidth}
\centerline{\includegraphics[scale=0.23]{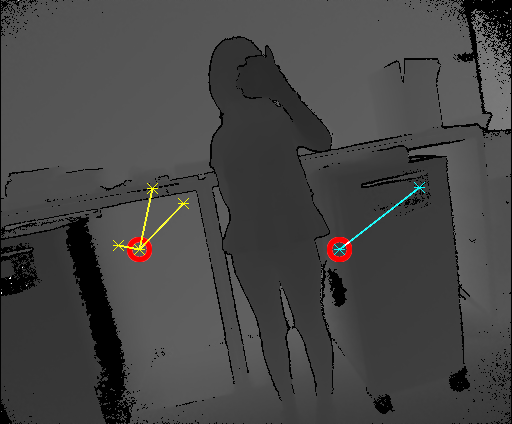}}
\end{minipage}
\begin{minipage}{0.49\linewidth}
\centerline{\includegraphics[scale=0.23]{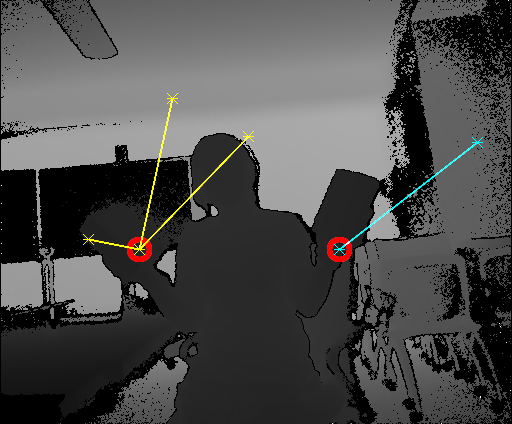}}
\end{minipage}
   \caption{Examples of the proposed feature representation. Each figure shows two different features where the same color line denotes the same feature. Each red circle represents a data point $\vx$. The length of each line denotes the magnitude of each offset ${\vu_i}/{\mD_{\vx}}$ considering the depth $\mD_{\vx}$ at the data point $\vx$. The end-point of each line shows each offset point $\vx+{\vu_i}/{\mD_{\vx}}$ used to compute the corresponding feature. The feature is computed by Eq. (\ref{eq:feature1}) where the training algorithm in~\ref{subsec:training} learns weights (coefficients) $w$, offsets $\vu$, and the number $n_f$ of data points.}\label{fig:Feature}
\end{center}\end{figure}

%% file: fig_feature_shape.tex
\begin{figure}[!t] \begin{center}
\begin{minipage}{0.325\linewidth}
\centerline{\includegraphics[scale=0.525]{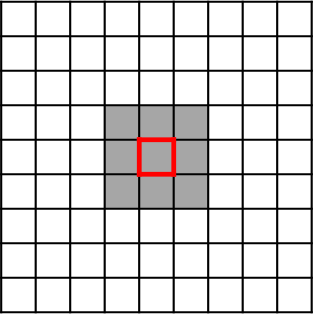}}
\centerline{\footnotesize} 
\end{minipage}
\begin{minipage}{0.325\linewidth}
\centerline{\includegraphics[scale=0.525]{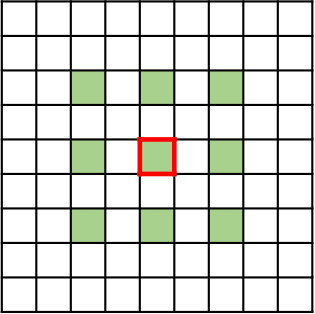}}
\centerline{\footnotesize (a) Feature in convolution layer} 
\end{minipage}
\begin{minipage}{0.325\linewidth}
\centerline{\includegraphics[scale=0.525]{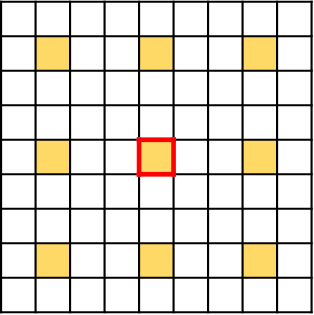}}
\centerline{\footnotesize} 
\end{minipage}

\vspace{0.25cm}
\begin{minipage}{0.325\linewidth}
\centerline{\includegraphics[scale=0.525]{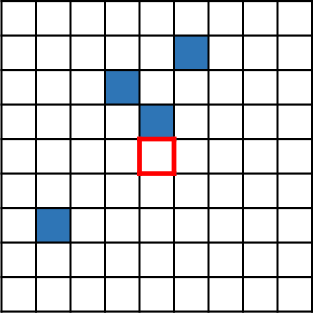}}
\centerline{\footnotesize} 
\end{minipage}
\begin{minipage}{0.325\linewidth}
\centerline{\includegraphics[scale=0.525]{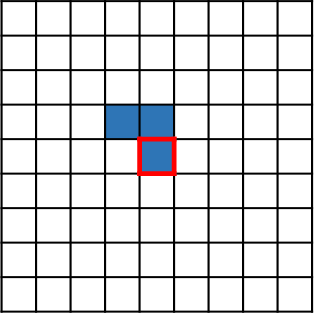}}
\centerline{\footnotesize (b) Proposed feature} 
\end{minipage}
\begin{minipage}{0.325\linewidth}
\centerline{\includegraphics[scale=0.525]{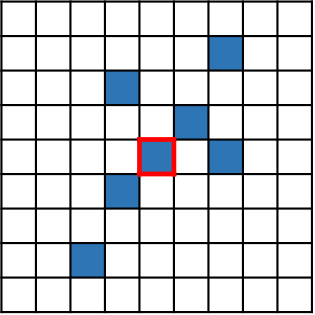}}
\centerline{\footnotesize} 
\end{minipage}
      \caption{Examples of the proposed representation and the representation of a convolutional layer in convolutional neural networks. The proposed representation is not constrained by a specific shape. Color squares denote offset points $\vx+{\vu}/{\mD_{\vx}}$ used to compute the feature, and the red bounding boxes represent the data points $\vx$.}
\label{fig:FeatureShape}
\end{center}\end{figure}

%% file: 3_3_random_forest.tex
\subsection{Random Forest} \label{subsec:RF}
\input{fig_RF_boosting}

Random forest consists of a set of decision trees as shown in Figs.~\ref{fig:Intro} and~\ref{fig:RF}. It is robust to noisy and irregular data because of the combination of multiple trees with varying features. It is also computationally less complex than typical neural networks since, in the inference procedure, each input data is processed using only small portion of the forest based on the conditions in ancestral nodes (see blue dotted lines in Figs.~\ref{fig:Intro} and~\ref{fig:RF}).

Each decision tree in random forest is composed of a root node, splitting nodes, and leaf nodes (see~\fref{fig:RF}). The input of each tree is the location $\vx \in \R^{2}$ of an input data and the corresponding input image and depth map ($\mX$, $\mD$). Given the input $(\vx, \mX, \mD)$ at a root node, the input data is classified to a child node based on the splitting criteria $f_n(\vx, \mX, \mD) \lessgtr \theta$ where $f_n(\cdot)$ extracts a learned feature at the node $n$. The classification to a descendant node is terminated when the input data reaches a leaf node. At the leaf node, the conditional probability $p(c|f_n(\vx, \mX, \mD) \lessgtr \theta \text{ for } \forall n \text{ until the leaf node})$ of being each class $c$ is learned in a training stage and is used in an inference stage. The leaf nodes are also learned in a training procedure. For more details about random forest, we refer readers to~\cite{RFBook, Breiman2001, Ho1995, RFKeypoint}.

\subsubsection{Training} \label{subsec:training}
In a training stage, the random forest learns a splitting criteria $f_n(\vx, \mX, \mD) \lessgtr \theta$ at each splitting node $n$ and a conditional probability distribution $p(c|f_n(\vx, \mX, \mD) \lessgtr \theta \text{ for } \forall n \text{ until the leaf node})$ of being each class $c$ at each leaf node. The splitting criteria at a node $n$ is denoted as follows:
\begin{equation}
f_n(\vx, \mX, \mD) = \sum_{i=1}^{n_f} w_i \mX_{\vx+{\vu_i}/{\mD_{\vx}}} \lessgtr \theta.
\label{eq:splitcriteria}
\end{equation}
In~(\ref{eq:splitcriteria}), one redundant parameter can be eliminated by dividing each side by $\theta$ while the data split remains equivalent.
\begin{equation}
\begin{split}
\sum_{i=1}^{n_f} \frac{w_i}{\theta} \mX_{\vx+{\vu_i}/{\mD_{\vx}}} & \lessgtr 1,  \\
\sum_{i=1}^{n_f} w'_i \mX_{\vx+{\vu_i}/{\mD_{\vx}}} & \lessgtr 1  \text{ where } w_i' = \frac{w_i}{\theta}.
\end{split}
\end{equation}
Thus, the training algorithm only needs to learn the parameters in the representation function $f(\cdot)$ while the splitting boundary is always 1. In the rest of this paper, a weight $w$ represents $w'$. 

In the following paragraphs, we introduce the strategy of selecting training data using bootstrap aggregating and boosting (see~\fref{fig:Boosting}). We then present the method of learning representations using particle swarm optimization. Finally, we describe the condition for determining leaf nodes and the process of learning conditional probability distribution.

\textbf{Bootstrap aggregating.}
Bootstrap aggregating (bagging) is to amalgamate multiple classifiers trained using randomly selected training data sets. This method improves robustness and accuracy by integrating multiple classifiers with variance caused by the randomly selected training data. Given the training data sets $\calX$ of images, the bagging algorithm selects random sets $\calX_i \subset \calX$ of images for the $i$-th tree. Then, the algorithm samples $n_d$ data points (pixels) for each class from the images $\mX \in \calX_i$. Thus, each tree is trained using $n_d \times n_c$ data points where $n_c$ is the total number of classes.

For hand segmentation, we sample the same number of data points for each image and for each class to consider diverse images equally regardless of the sizes of hands, etc. For semantic segmentation, since all objects do not appear on all the images, we only constrain the same number of data points for each class.

\textbf{Boosting.} Boosting is to combine a set of weak classifiers to form a stronger classifier. Especially, adaptive boosting is learning weak classifiers based on the obtained result by applying previously trained weak classifiers~\cite{Freund}. We apply adaptive boosting at each set of trees by adjusting the sampling of training data based on the segmentation result at the stage.

As explained in the bagging, a random set $\calX_i \subset \calX$ of images is selected for the $i$-th tree $T_i$. Then, $n_d$ data points $\vx$ are sampled for each class $c$. The proposed boosting algorithm adjusts this sampling of data points $\vx$ for each set of trees. For the first set of trees, data points $\vx$ are sampled randomly with uniform distribution for each class. Thus, the probability $p_s(\cdot)$ of being sampled for a data point $\vx$ of a class $c$ is as follows:
\begin{equation}
p_s(\vx) = \frac{n_d}{n_{d_c} n_{\mX}} 
\end{equation}
where $n_{d_c}$ is the number of the data points of a class $c$ in the image $\mX$ for hand segmentation and in the set $\calX$ of images for semantic segmentation, respectively. $\mX$ is the image that $\vx$ belongs to. $n_{\mX}$ is 1 for semantic segmentation and is the number of images in $\calX$ for hand segmentation. 

After training the first set of trees, the sampling probability $p_s(\cdot)$ is updated to train the second set of trees more effectively. It is achieved by adjusting $p_s(\cdot)$ to sample more data points with higher errors. Hence, $p_s(\cdot)$ is increased for the data points with a high error and is decreased for the data with a low error. Given the learned set of trees $\calT$ until the current iteration, an inference is processed to estimate the probability $p_c(c|\vx, \mX, \mD)$ of being a target class $c$ for the data in the training dataset $\calX$. Then, $p_s(\cdot)$ is updated as follows:
\begin{equation}
p_s(\vx) = 1 - p_c(c|\vx, \mX, \mD).
\end{equation}
By using the data points sampled based on the updated $p_s(\cdot)$, the next set of trees is trained. $p_s(\cdot)$ is adjusted repeatedly at each iteration until the training terminates.

\fref{fig:Boosting} shows a visual example of boosting and bootstrap aggregating. The first row and the second row show the first iteration and the second iteration of boosting, respectively. In the first iteration, $p_s(\cdot)$ in (b) is the same for all data points of the same class. Thus, the sampled data points in (c) are distributed uniformly. In the second iteration, $p_s(\cdot)$ in (f) varies for each data point depending on the inferred probability in (d) that is obtained using the first set of trees. Consequently, the sampled data points in (g) are distributed based on $p_s(\cdot)$ in (f).

\textbf{Particle swarm optimization.} Particle swarm optimization (PSO) is applied to learn a feature representation that splits the data points of different classes into separate child nodes at each splitting node. PSO is selected to find a more optimal solution in a high-dimensional parameter space ($\mathbb{R}^{3 \times n_f} \cdot \mathbb{Z}^{1}$ if $\mX$ is a depth map, and additional $\mathbb{Z}^{n_f}$ if $\mX$ is a color image). 

Assuming $\mX$ is a depth map, a representation parameter vector $\vp \in \mathbb{R}^{3 \times n_f} \cdot \mathbb{Z}^{1}$ consists of the number $n_f \in \mathbb{Z}^+$ of data points and a weight $w \in \mathbb{R}$ and an offset parameter vector $\vu \in \mathbb{R}^2$ for each data point. Thus, the total number of parameters is $3 \times n_f + 1$. To limit the solution space ($\mathbb{R}^{3 \times n_f} \cdot \mathbb{Z}^{1}$), the maximum number of data points $n_f$ is chosen as 9 which is equivalent to the number of data points in a filter (kernel) with the size of $3\times3$. It is also the most common size of a kernel in convolutional neural networks.

In the first iteration, the algorithm generates 100 offset candidates $\vu$ and 100 weight candidates $w$ where the candidates are sampled from the uniform distribution. Then, the algorithm tries the combinations of the 9 different numbers $n_f$ of data points, 100 offset candidates $\vu$, and 100 weight candidates $w$, totaling 90,000 particles (candidates). By applying the particles, the optimization algorithm learns the global best state $\vq_g$ that is the best solution among the entire particles. To simplify and expedite training, the number $n_f$ of data points is decided as the number of data points of the global best state $\vq_g$ in the first iteration. Also, among 10,000 particles, 100 particles are chosen by selecting the best weight candidate for each offset candidate. It is feasible since 10,000 particles are generated by the combinations of 100 offset candidates $\vu$ and 100 weight candidates $w$.

From the second iteration, the 100 particles are used to find the optimal solution. Let $\vq \in \mathbb{R}^{3 \times n_f}$ represents each particle consisting of weights $w$ and offsets $\vu$. 
At an iteration $t$, the particles $\vq^t$ are first updated using the particles $\vq^{t-1}$, the personal best states $\vq_p^{t-1}$, and the global best state $\vq_g^{t-1}$ at the previous iteration $t-1$. The personal best state $\vq_p^t$ is the best state of each particle until the current iteration $t$. The global best state $\vq_g^t$ is the best state among all the particles until the current iteration $t$. The particles at an iteration $t$ are updated as follows:
\begin{equation}
\begin{split}
& \vq_{p,i}^{t-1} = \{\vq_i^{\ttt} | \ttt = \argmin{t} L(q_i^{t}) \}, \\
& \vq_g^{t-1} = \{ \vq_{\ti}^{\ttt} | (\ti, \ttt) = \argmin{i, t} L(q_{i}^{t}) \}, \\
& \vq_i^{t} = \vq_i^{t-1} + \alpha_p (\vq_{p,i}^{t-1} - \vq_i^{t-1}) + \alpha_g (\vq_g^{t-1} - \vq_i^{t-1}) 
\end{split}
\end{equation}
where $i$ denotes the index for each particle. $L(\cdot)$ is the objective function in (\ref{eq:loss}). $\alpha_p$ and $\alpha_g$ are the weights towards the personal best state $\vq_p$ and the global best state $\vq_g$. Both parameters ($\alpha_p$ and $\alpha_g$) are randomly generated from the Normal distribution $\calN(\mu, \sigma^2)$ as follows:
\begin{equation}
\begin{split}
&\widetilde{\alpha} \sim \calN(1.0, 0.25), \\
&\alpha = \max(0, \widetilde{\alpha})
\end{split}
\end{equation}
where $\mu$ and $\sigma$ represent a mean and a standard deviation. The iteration of PSO is terminated when the loss $L(\cdot)$ in (\ref{eq:loss}) is not decreased at each iteration or after maximum 100 iterations.

The objective function $L(\cdot)$ consists of a term for classification loss and two terms for regularization. The classification loss is to evaluate a representation's ability of separating the data points of different classes to separate child nodes. The regularization is to prefer smaller weights $w$ and the smaller number $n_f$ of data points.
\begin{equation}
\begin{split}
L(\vp) = & \underbrace{C(\vp)}_{\text{classification loss}} + \underbrace{\lambda_w \sum_{i=1}^{n_f} {w_i}^2 + \lambda_{n_f} n_f}_{\text{regularization}} \\
 = & \underbrace{- \sum_{h}\sum_{c}\frac{n(h)}{\sum_h n(h)} p(c|h) \log{p(c|h)}}_{\text{classification loss}} \\ 
& + \underbrace{\lambda_w \sum_{i=1}^{n_f} {w_i}^2 + \lambda_{n_f} n_f}_{\text{regularization}}
\label{eq:loss}
\end{split}
\end{equation}
where $h$ is an index for a child node (e.g. left child, right child); $c$ is an index for a class; $n(h)$ denotes the number of data points in a child node $h$; $p(c|h)$ is the probability of being the class $c$ at the node $h$; $\lambda_w$ and $\lambda_{n_f}$ are weights for each regularization term. 


After learning an optimal representation at a splitting node, the data is split into child nodes. If the child node does not meet the condition for becoming a leaf node, learning representation and splitting to child nodes are repeated. Otherwise, splitting is terminated, and a leaf node is formed.


\textbf{Leaf node.} A leaf node is formed based on the following criterias: (1) the maximum depth of a tree, (2) the probability distribution $p(c|h)$, and (3) the number of training data $\vx$ at the node. In details, a leaf node is generated if (1) the current depth of a tree is deeper than the maximum depth; (2) the probability at the node is considerably confident for a class; (3) the number of remaining training data is too small. When a leaf node is formed, the conditional probability is stored for inference processing. The conditional probability $p(c|f_n(\vx, \mX, \mD) \lessgtr \theta \text{ for } \forall n \text{ until the leaf node})$ is computed using the number of data points for each class at the leaf node $h$.
\begin{equation}
\begin{split}
&p(c|f_n(\vx, \mX, \mD) \lessgtr \theta \text{ for } \forall n \text{ until the leaf node}) \\
&= p(c|h)  = \frac{n(h,c)}{\sum_{c=1}^{n_c} n(h,c)}
\end{split}
\end{equation}
where $n(h,c)$ represents the number of data points for each class $c$ at the leaf node $h$.

\subsubsection{Inference} \label{subsec:inference}
Given the trained random forest $\calT$, each data point $\vx$ on an image $\mX$ is classified to child nodes using each tree until it reaches a leaf node. When it reaches a leaf node using each tree $T_i$, the learned conditional probability distribution $p_{T_i}(c|f_n(\vx, \mX, \mD) \lessgtr \theta \text{ for } \forall n \text{ until the leaf node})$ of being each class $c$ at the leaf node is loaded. Then, the conditional probability distributions from the entire trees $T_i \in \calT$ are averaged to estimate the inferred probability $p_c(c|\vx)$ of the data point $\vx$ being a class $c$. 
\begin{equation}
\begin{split}
p_c(c|\vx) = \frac{1}{n_t} \sum_{i=1}^{n_t} p_{T_i}(c|f_n(\vx, \mX, \mD) \lessgtr \theta \\
\text{ for } \forall n \text{ until the node})
\end{split}
\end{equation}
where $n_t$ is the number of trees in the random forest $\calT$.



%% file: fig_RF_boosting.tex
\begin{figure*}[!t] \begin{center}
\begin{minipage}{0.20\linewidth}
\centerline{\includegraphics[scale=0.20]{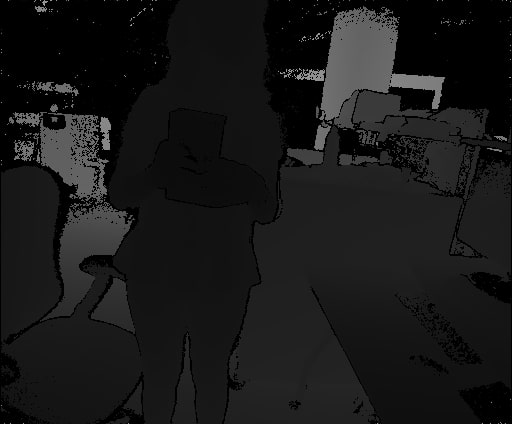}}
\centerline{\footnotesize (a) Depth map} 
\end{minipage}
\begin{minipage}{0.20\linewidth}
\centerline{\includegraphics[scale=0.20]{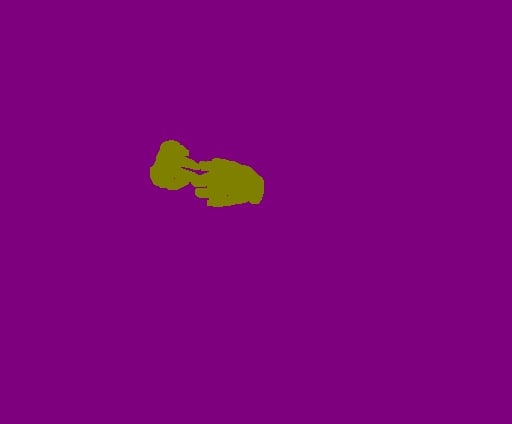}}
\centerline{\footnotesize (b) Sampling probability} 
\end{minipage}
\begin{minipage}{0.20\linewidth}
\centerline{\includegraphics[scale=0.20]{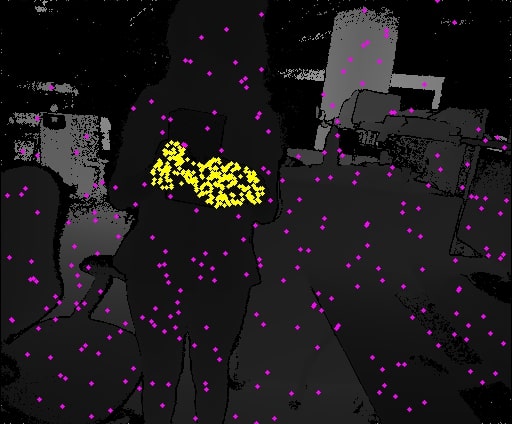}}
\centerline{\footnotesize (c) Sampled data} 
\end{minipage}
\begin{minipage}{0.20\linewidth}
\centerline{\includegraphics[scale=0.20]{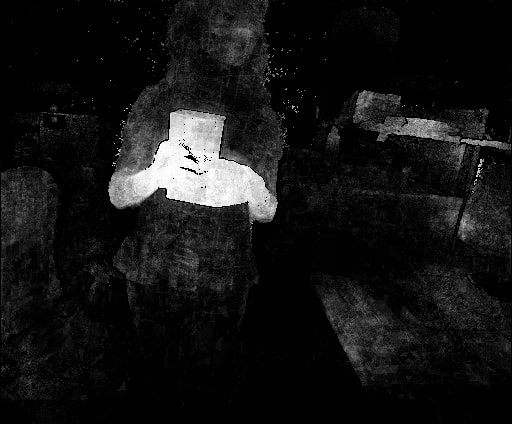}}
\centerline{\footnotesize (d) Inferred probability} 
\end{minipage}

\vspace{0.1cm}
\begin{minipage}{0.20\linewidth}
\centerline{\includegraphics[scale=0.20]{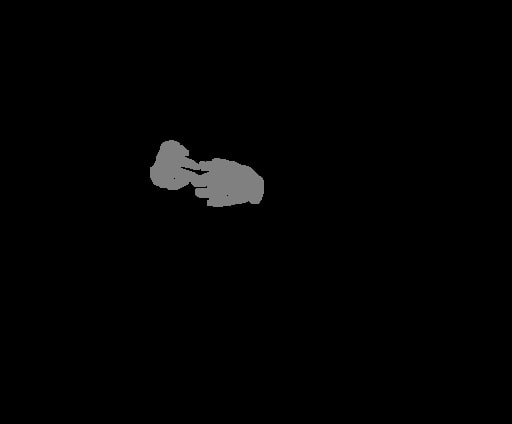}}
\centerline{\footnotesize (e) Ground truth label} 
\end{minipage}
\begin{minipage}{0.20\linewidth}
\centerline{\includegraphics[scale=0.20]{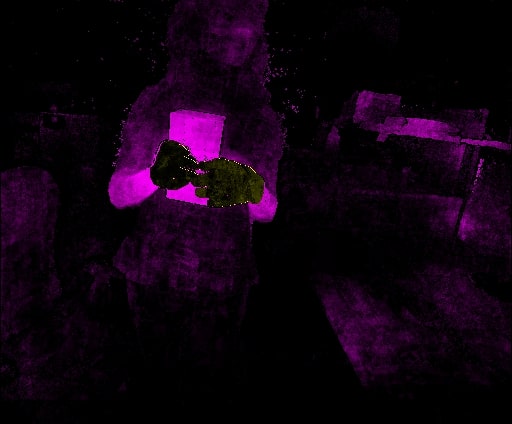}}
\centerline{\footnotesize (f) Sampling probability} 
\end{minipage}
\begin{minipage}{0.20\linewidth}
\centerline{\includegraphics[scale=0.20]{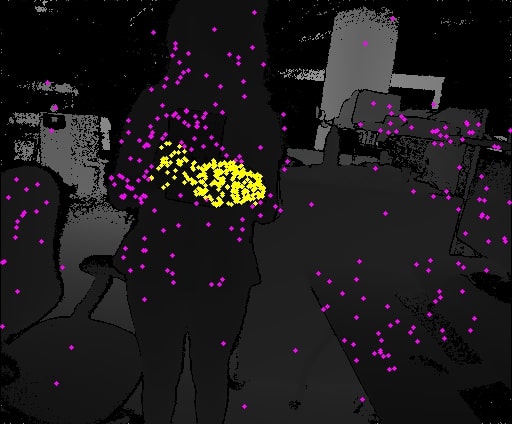}}
\centerline{\footnotesize (g) Sampled data} 
\end{minipage}
\begin{minipage}{0.20\linewidth}
\centerline{\includegraphics[scale=0.20]{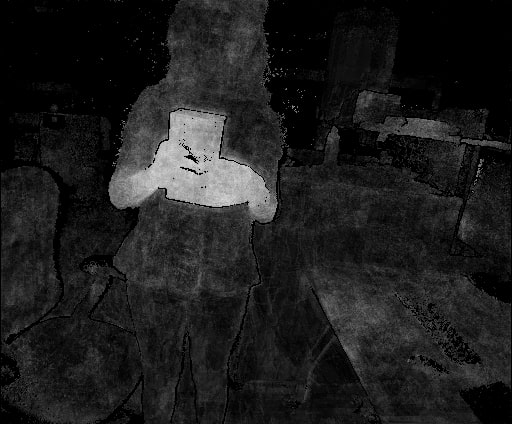}}
\centerline{\footnotesize (h) Inferred probability} 
\end{minipage}
   \caption{Example of boosting and bootstrap aggregating. (a) Depth map. (b) Sampling probability $p_s(\cdot)$ in the first iteration of boosting. (c) Sampled data points in the first iteration. (d) Inferred probability using the first set of learned trees. (e) Ground truth label. Gray and black color denote hand and non-hand class, respectively. (f) Sampling probability $p_s(\cdot)$ in the second iteration of boosting. (g) Sampled data points in the second iteration. (h) Inferred probability using the first and the second set of trees. Magenta and yellow color represent non-hand and hand class, respectively. Brighter color in (b) and (f) denotes higher sampling probability $p_s(\cdot)$. Brighter color in (d) and (h) represents the higher probability of being hand class. The number $n_d$ of sampled data points is 250 for each class in this example.}
\label{fig:Boosting}
\end{center}\end{figure*}

%% file: 3_4_filtering.tex
\textbf{Modified bilateral filter.}
Since the probability $p_c(c|\vx)$ is predicted for each pixel independently, the probability can be stabilized considering nearby predictions~\cite{CRF, kangglobalsip}. In this paper, we apply simple modified bilateral filter~\cite{tomasi, kangglobalsip} that processes weighted averaging of the probabilities of the data points in close distance and similar intensity on the input $\mX$. 
The filtering is defined as follows:
\begin{equation}
\begin{split}
\ptilde_c(c|\vx) = \frac{1}{N} \sum_{\vx_i \in \Omega}  & g_r( \abs{\mX_{\vx_i} - \mX_{\vx}})    g_s(\norm{\vx_i - \vx}) p_c(c|\vx_i)
\end{split}
\end{equation}
where $\Omega$ is the set of pixels within the filter's radius and the input difference; $N$ is the normalization term.
\begin{equation}
N = \sum_{\vx_i \in \Omega}  g_r( \abs{\mX_{\vx_i} - \mX_{\vx}}) g_s(\norm{\vx_i - \vx}).
\end{equation}
$g_r(\cdot)$ and $g_s(\cdot)$ are the Gaussian functions for an input difference $r$ and a spatial distance $s$, respectively.
\begin{equation}
g_r(r) = \exp \bigg(-\frac{r^2}{2\sigma_r^2}\bigg), \quad g_s(s) = \exp \bigg(-\frac{s^2}{2\sigma_s^2} \bigg).
\end{equation}
By experimenting on the validation dataset, the maximum color and depth difference to consider are 100 and 400$mm$, respectively. Both the standard deviations ($\sigma_r$ and $\sigma_s$) are 100.

\begin{figure}[!t]
\centering
    \includegraphics[width=0.40\textwidth]{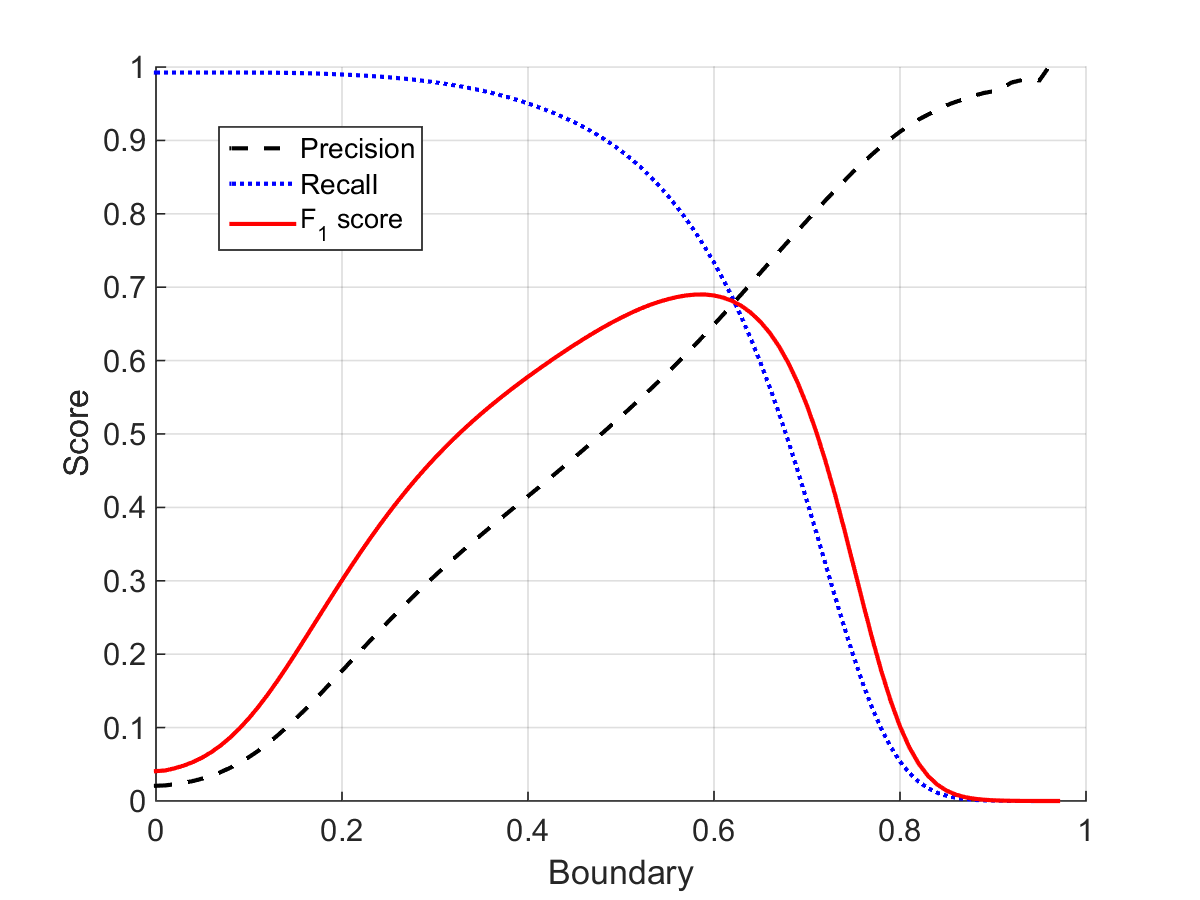}
   \caption{Scores depending on decision boundary. The scores are computed on the validation set of hand segmentation data using the proposed method (30 trees) without the bilateral filtering.}
\label{fig:classificationDecision}
\end{figure}

\input{tbl_result_hand}

\textbf{Decision boundary.}
Given the filtered probability $\ptilde_c(c|\vx)$, a decision boundary needs to be determined for classification. Although the most common method is choosing the class $c$ with the highest probability $\ptilde_c(c|\vx)$, it is not guaranteed to be the best solution. Thus, for hand segmentation, the possible boundaries are tested with the step size of 0.01 using the validation dataset. \fref{fig:classificationDecision} shows the $F_1$ score, precision, and recall on the validation dataset depending on the decision boundary. The scores are computed using the proposed method with 30 trees and without the bilateral filtering. The best $F_1$ score is achieved at the decision boundary of 0.59. For semantic segmentation, the class with the highest probability is selected considering the high complexity caused by the high dimensional decision boundary.

%% file: tbl_result_hand.tex
\begin{table*}[!t]
\caption{The quantitative results of the HOI dataset.}
\label{tab:result}
\centering
\renewcommand{\arraystretch}{1.1}
\begin{tabu} to 0.96\textwidth { c|X[c,m]|X[c,m]|X[c,m]|X[c,m]|X[c,m]|X[c,m]|X[c,m]}
\hline
\multicolumn{4}{c|}{Method} & \multicolumn{3}{c|}{Score} & Time ($ms$) \\
\hline
\multirow{2}{*}{Method} & \multirow{2}{*}{\# of trees} & Decision & Bilateral & \multirow{2}{*}{Precision} & \multirow{2}{*}{Recall} & \multirow{2}{*}{$F_1$ score} & \multirow{2}{*}{GTX 770} \\
 &  & boundary & filter &  &  &  &  \\
\hline \hline
\multirow{2}{*}{RF~\cite{shottoncvpr, tompson}}			& 20 & 0.50	& -	& 37.9 & 91.9 & 53.7 & 10 \\
\cline{2-8}
														& 30 & 0.50	& -	& 37.9 & 92.1 & 53.7 & 14 \\
\hline
\multirow{2}{*}{RF~\cite{shottoncvpr, tompson} + DB adjustment}	& 20 & 0.79	& -	& 54.9 & 72.7 & 62.5 & 10 \\
\cline{2-8}
														& 30 & 0.79	& -	& 54.8 & 73.0 & 62.6 & 14 \\
\hline
FCN-32s~\cite{longcvpr}									& -	& -		& -	& 70.0 & 68.6 & 69.3 & 376\\
FCN-16s~\cite{longcvpr}									& -	& -		& -	& 68.0 & 72.2 & 70.1 & 376 \\
FCN-8s~\cite{longcvpr}									& -	& -		& -	& 70.4 & 74.4 & 72.3 & 377 \\
Frontend~\cite{YuKoltun2016}								& -	& -		& -	& 72.4 & 70.2 & 71.3 & 718 \\
\hline 
\multirow{5}{*}{Proposed}								& 20 & 0.50	& -				& 50.7 & 89.4 & 64.7 & 28 \\
														& 20 & 0.59	& -				& 62.2 & 76.3 & 68.5 & 28 \\
\cline{2-8}
														& 30 & 0.50	& -				& 53.4 & 88.4 & 66.5 & 39 \\
														& 30 & 0.59	& - 				& 64.7 & 75.7 & 69.8 & 39 \\
														& 30 & 0.59	& $5\times5$ 	& 65.3 & 76.0 & 70.3 & 41 \\
\hline
\end{tabu}
\end{table*}

%% file: 4_experiments_results.tex
\input{fig_result_hand}

\input{tbl_analysis_depth_hand}

\begin{figure}[!t]
\centering
	\includegraphics[width=0.37\textwidth]{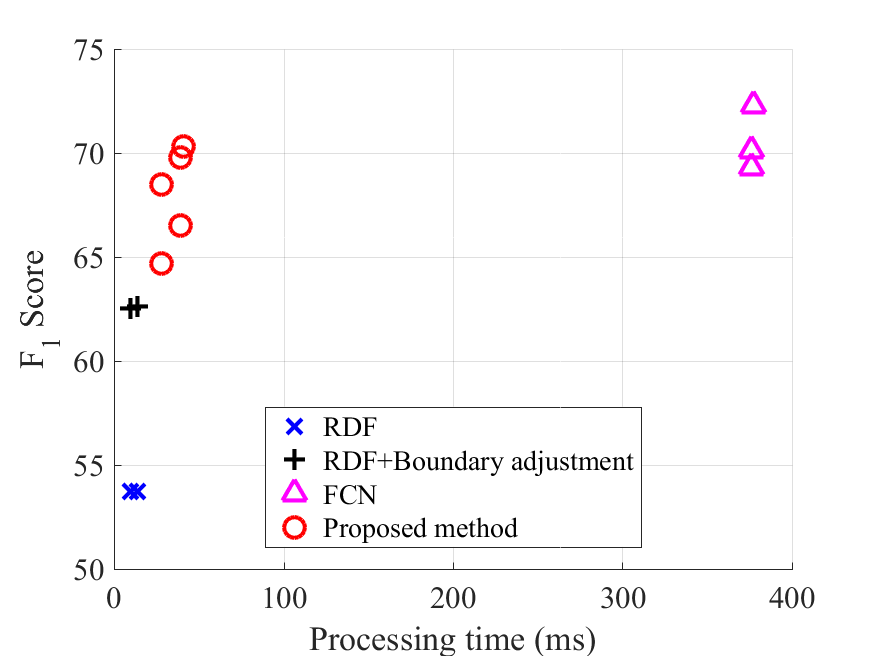}
   \caption{Analysis in accuracy and efficiency.}
\label{fig:accuracyEfficiency}
\end{figure}

\section{Experiments and Results} \label{sec:result}
The proposed method is applied to three applications: road scene semantic segmentation, hand segmentation for hand-object interaction, and indoor scene semantic segmentation. The experimental results demonstrate that the proposed method outperforms typical random forest by learning unconstrained representations using particle swarm optimization and processes efficiently comparing to neural network-based methods.

For comparison, we report mean intersection over union (IU) per category and per class for road scene semantic segmentation dataset and precision, recall, and $F_1$ score for hand segmentation for hand-object interaction. Let $n_{ij}$ be the number of pixels that belong to the class $i$ and are predicted to the class $j$, and $n_c$ be the total number of classes or categories. 
\begin{equation}
\begin{split}
&\text{IU} = \frac{1}{n_c} \sum\limits_i \bigg( \frac{n_{ii}}{\sum\limits_j n_{ij} + \sum\limits_j n_{ji} - n_{ii}} \bigg),         \\
&\text{Precision} = \frac{n_{11}}{n_{11} + n_{01}},    \\
&\text{Recall} = \frac{n_{11}}{n_{11} + n_{10}},          \\
&F_1 = \frac{2 n_{11}}{2 n_{11} + n_{01} + n_{10}},
\end{split}
\end{equation}

         



\noindent where for hand segmentation, classes 1 and 0 denote ``hand'' and ``others'', respectively.

For quantitative comparison of efficiency, we measure processing time using a machine with Intel i7-4790K CPU with 4.00GHz, 16.0GB RAM, and NVIDIA GeForce GTX 770 for hand segmentation and the same machine with NVIDIA Tesla K40c for semantic segmentation.

\subsection{Hand-Object Interaction~(HOI)} \label{sec:HOI}
\subsubsection{Dataset}  
We experiment using the publicly available HOI dataset~\cite{kangtmm} which consists of 27,525 pairs of depth maps and ground truth labels. The dataset was collected from 6 people~(3 males and 3 females) interacting with 21 different objects using Microsoft Kinect v2 camera. The dataset also includes the cases of one hand and both hands in a scene. We follow standard dataset split (19,470 pairs for training, 2,706 pairs for validation, and 5,349 pairs for testing).

\subsubsection{Experiments}  
We trained 30 trees using the proposed method. Each tree is trained with 2,000 pairs of depth maps and ground truth labels and 500 data points from each image. After training every 10 trees, we applied boosting by computing errors and by sampling based on the errors. The condition for becoming a leaf node was 0.99 for class probability, 0.0001 for remaining data portion, and 25 for maximum depth.

\subsubsection{Result} \label{subsec:result}
The quantitative results and the qualitative results are shown in~\tref{tab:result} and~\fref{fig:result}, respectively. The proposed method achieves about 31\% and 12\% relative improvement in $F_1$ score comparing to the typical RF-based method~\cite{shottoncvpr, tompson} and its combination with the decision boundary adjustment in~\sref{subsec:inference}. Comparing to the deep learning-based methods~\cite{longcvpr, YuKoltun2016}, it achieves quite competitive results (3\% lower than the best method) in $F_1$ score. Also, the processing time of the proposed method is about 9 times faster than those of the deep learning-based methods~\cite{longcvpr, YuKoltun2016}. \fref{fig:accuracyEfficiency} shows the analysis of the methods in accuracy and efficiency. In~\tref{tab:maximumDepthHOI}, we show empirical results of changing the maximum depth in the proposed random forest. In this analysis, the filter-size of bilateral filtering is 5$\times$5.


\input{tbl_result_cityscapes}

\subsection{Semantic Segmentation~(Cityscapes)}
\subsubsection{Dataset}  
The Cityscapes dataset contains the images of urban street scenes~\cite{cityscapes}. The dataset consists of 5,000 finely annotated images and 19,998 sparsely annotated images. We train models for the standard 19 classes problem using the standard data separation of 2,975 finely annotated images and 19,998 sparsely annotated images for training, 500 images for validation, and 1,525 images for testing.

\subsubsection{Experiments} \label{cityscapes_experiments}
We trained five trees using the proposed framework. Each tree is trained with [12,974, 12,973, 7,658, 7,656, 7,658] pairs of images and ground truth labels. The first two trees are trained using the entire finely annotated images and the half of sparsely annotated images. The last three trees are trained with 1/3 of entire training data sets. The selection of the number of images is based on both experimental and intuitive choice. After training each tree, we applied boosting by sampling the set of data points for the next tree based on the current predictions. The condition for becoming a leaf node is 0.99 for class probability, [0.000001 or 0.000002] for remaining data portion, and 25 for maximum depth.

Although predicting using random forest is a computationally and memory efficient process, training large forest with a huge amount of data is computationally expensive. It is especially time-consuming during training procedure since each node needs to be optimized conditioned on ancestor nodes where the number of nodes can be up to 67,108,863 considering the maximum depth of 25. Hence, we altered the training algorithm to reduce learning time.

\input{fig_result_cityscapes}

One of the most time-consuming processes is loading training images repeatedly. We used over 7,656 images to train each tree where the original resolution of each image is 2048$\times$1024. The required memory to hold 7,656 color images is 7,656$\times$2,048$\times$1,024$\times$3 bytes (44.9 GB) which are not accessible in modern single GPU. One option is loading partial sets of images multiple times at each node. However, it is a considerably time-consuming process since loading 1,000 images from a hard disk drive takes 51 seconds and repeating at 1,000 nodes demands 14 hours. Hence, we propose to decrease the resolution of images to hold the entire set of training data in memory after loading once for a tree. Specifically, we resized images to 512$\times$256 so that the required memory to hold 12,974 color images is 4.8 GB.  Given 11 GB in modern single GPU, the rest of memory is utilized to process learning algorithms.

Using the remaining memory, we were able to load and process 103,792/382,900 samples for 12,974/7,658 color images where each sample consists of frame index, $\vx$, $\mD_{\vx}$, and label. However, this number of samples is too small since the average number of data points at depth 15 becomes 3.2 (103,792 / $2^{15}$) and 11.7 (382,900 / $2^{15}$). Hence, we employ multiple sets of data points (specifically, 64 sets) so that the average number at depth 15 becomes 202.7 and 747.9 data points. We start training with 64 sets at a root node and merge the sets at a deeper node since the number of data points in a set decreases as the data splits to child nodes. While multiple sets are employed to provide enough data, we use a single set (possibly, merged from multiple sets) among them to train a splitting node in order to reduce computation and data transfer time.

\input{tbl_analysis_cityscapes}

\input{tbl_analysis_PSO}

\input{tbl_analysis_resize}

\subsubsection{Results}  
We compare the proposed method to deep learning-based methods~\cite{longcvpr, longpami, Zheng, enet, segnet, OCNet, CCNet, refinenet, icnet, pspnet} in~\tref{tab:cityscapesResult} and~\fref{fig:cityscapesResult}. We also show the results of the proposed method with various conditions on the validation set in~\tref{tab:cityscapesAnalysis}. We report mean IU per category and per class, processing time, and memory usage. Since optimizing post-processing is beyond the scope of this paper, we report processing time and memory usage excluding the demand for bilateral filtering. The overall results show that the proposed method processes each image using small computation and memory resources while achieving meaningful precision. It demonstrates that the proposed method can be applied for real-time semantic segmentation. It can also be employed in a low-end GPU or embedded system that demand small power and memory consumption.



We show the results of applying the same random forest model to the inputs of varying resolutions using the validation set in~\tref{tab:cityscapesAnalysis}. The results demonstrate that the proposed method is robust to scaled input images by simply adjusting the sparsities in the learned representations. It is feasible since the sparsity in the proposed representation is defined in floating-point precision. \tref{tab:cityscapesResize} shows the results of applying the FCN model trained with the input resolution of 2048$\times$1024 to the inputs of different resolutions using the validation set. The accuracy degrades significantly since deep learning-based methods utilize integer-point precision (mostly, 1) for sparsity. 

\subsubsection{Analysis}  
We demonstrate the effectiveness of the proposed unconstrained representation and the particle swarm optimization comparing to typical convolution filter and random search~\cite{shottoncvpr, tompson, kangglobalsip} in~\tref{tab:pso}. By comparing the unconstrained representation and the convolution filter, the decision tree trained using the unconstrained representation achieves higher accuracy than the tree with the typical convolution filter even with a smaller number of nodes, shorter processing time, and lower GPU memory usage in both optimization methods. For the typical convolution filter, 3$\times$3$\times$3 convolution filters are used to take inputs from 3$\times$3 spatial regions and entire color channels (total 27 parameters). The analysis between particle swarm optimization and random search shows that the model trained with particle swarm optimization outperforms the other model optimized with random search while complexity (the number of nodes, processing time, and memory usage) is similar. It verifies that more optimal representation can be learned by using the unconstrained representation and the particle swarm optimization. In these analyses, the resolutions of input images were 512$\times$256 in training as explained in~\sref{cityscapes_experiments}.


In~\tref{tab:hyperparameters}, we show empirical results to decide hyperparameters for particle initialization in particle swarm optimization using the validation set. The hyperparameters are to set the ranges for the uniform distributions described in~\sref{subsec:training}. Although evolved particles can move outside of the initial boundary, initializing particles in a desirable range is important to search solution space efficiently using the limited number of particles and computational resources.

\input{tbl_analysis_hyperparameter}
\input{tbl_analysis_depth}

In~\tref{tab:maximumDepthCityscapes}, we show experimental analysis of adjusting the maximum depth in the proposed method. In this analysis, the input resolution and the filter-size of bilateral filtering are 512$\times$256 and 19$\times$19, respectively. The effects of varying the maximum depth in different datasets can be observed using Tables~\ref{tab:maximumDepthHOI} and~\ref{tab:maximumDepthCityscapes}. 

We present per-category accuracy in~\tref{tab:perCategory}. This analysis shows that the result of the proposed method is competitive for the categories of flat, sky, and nature comparing to CNN-based methods. However, the accuracy difference is greater for the categories of object and human that include relatively small and sparse things such as traffic light, traffic sign, and rider (also see~\fref{fig:cityscapesResult2}). We believe it is greater for the categories because of image resolution and random sampling. First, we train and test using images with 512$\times$256 resolution because of memory limit during training. This scaling makes small objects even tinier. Consequently, it is challenging to classify them in pixel-level during testing. Moreover, because of randomly sampling from resized images without data augmentation, the variety of data points for small object classes is limited. Accordingly, the trained forest has limited generality for new data.

\input{tbl_analysis_category}
\input{fig_result_cityscapes2}

While the proposed method has advantages in computational complexity, memory demand, and ability to control complexity by adjusting the number of trees and the input resolution, the accuracy of the proposed method is limited comparing to deep learning-based methods. We discuss three possible explanations. First of all, while deep neural networks can describe nonlinear representations by using a hierarchical structure with nonlinear activation functions, the representation of the proposed method is a linear representation of the original input image at the root node (see~\eref{eq:feature1}). Second, while deep neural networks are usually trained end-to-end by using back-propagation, the proposed method is trained in-order from a parent node to child nodes. As all the nodes are trained only once in-order, overall optimization might be relatively distant from global optima. Lastly, as Tables~\ref{tab:maximumDepthHOI} and \ref{tab:maximumDepthCityscapes} show, increasing the size of a tree can improve accuracy at the cost of processing time and memory usage. However, training a deeper tree is quite computationally expensive since the number of nodes increases exponentially as the depth increases.

\subsection{Semantic Segmentation~(NYUDv2)}
\subsubsection{Dataset}
The NYUDv2 dataset consists of 1,449 pairs of RGB-D images including various indoor scenes with pixel-wise annotations~\cite{nyudv2}. The pixel-wise annotations were coalesced into 40 dominant object categories by Gupta \etal~\cite{gupta2013}. We experimented with this 40 classes problem using the standard data separation~\cite{nyudv2, gupta2013} of 795 training images and 654 testing images. 

\subsubsection{Experiments}
We trained a decision tree using the proposed framework. To increase the variety of training data, the training images are augmented to total 5,565 images by randomly scaling, cropping, and altering color information. For scaling, a random number is sampled from a uniform distribution in [0.7, 1.3). If the scale is greater than one, the enlarged image is randomly cropped to an image with the original size. Concerning color alteration, images in RGB space are first transformed to images in HSV space. Then, values in H, S, and V space are scaled with a random number drawn from uniform distributions in [0.9, 1.1), [0.666, 1.5), and [0.666, 1.5), respectively. For training, we sampled 1,024 sets of 100,170 samples. The condition for becoming a leaf node is 0.9 for class probability, [0.000001] for the remaining data portion, and 25 for maximum depth.

\input{tbl_result_nyudv2}

\subsubsection{Result}
The quantitative results of an ablation study are shown in~\tref{tab:nyudv2Result}. Although the pixel accuracy of the proposed method is lower than deep neural network-based methods, the processing time of the proposed method is about 30 times faster than the other methods~\cite{longcvpr, longpami, YuKoltun2016, Chen16}.

%% file: fig_result_hand.tex
\begin{figure*}[!t] \begin{center}
\begin{minipage}{0.18\linewidth}
\centerline{\includegraphics[scale=0.13]{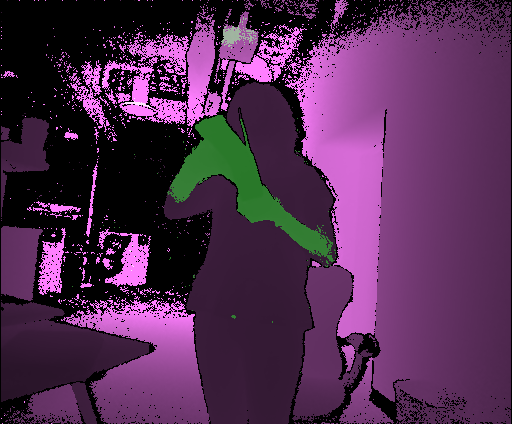}}
\end{minipage}
\begin{minipage}{0.18\linewidth}
\centerline{\includegraphics[scale=0.13]{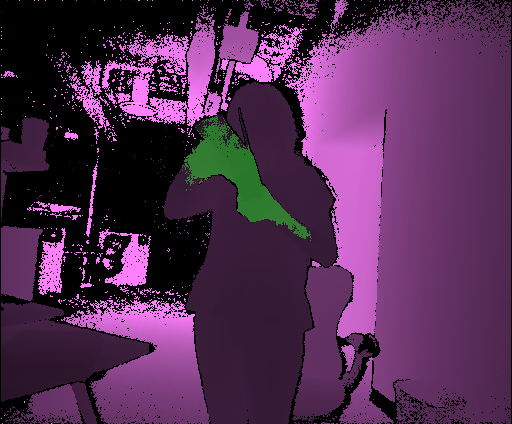}}
\end{minipage}
\begin{minipage}{0.18\linewidth}
\centerline{\includegraphics[scale=0.13]{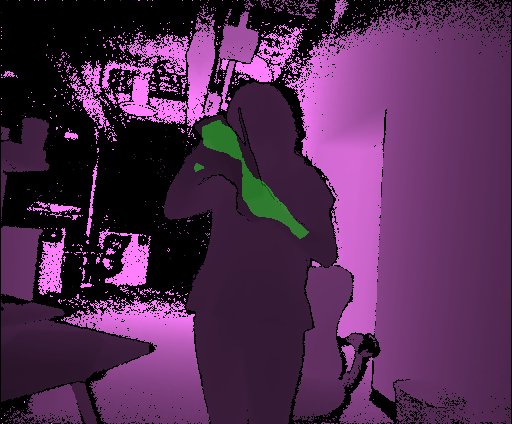}}
\end{minipage}
\begin{minipage}{0.18\linewidth}
\centerline{\includegraphics[scale=0.13]{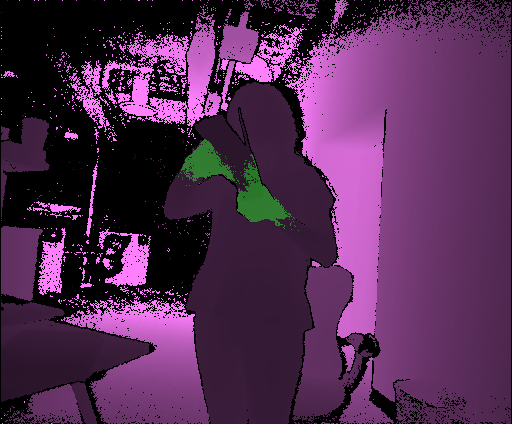}}
\end{minipage}
\begin{minipage}{0.18\linewidth}
\centerline{\includegraphics[scale=0.13]{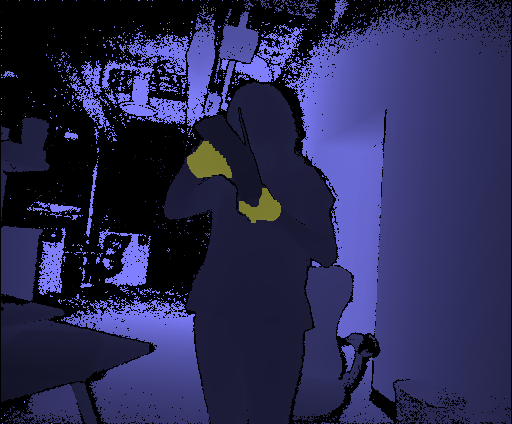}}
\end{minipage}
\\
\vspace{0.1cm}
\begin{minipage}{0.18\linewidth}
\centerline{\includegraphics[scale=0.26]{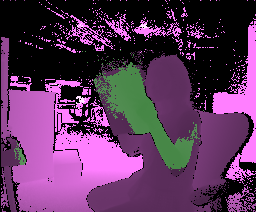}}
\end{minipage}
\begin{minipage}{0.18\linewidth}
\centerline{\includegraphics[scale=0.26]{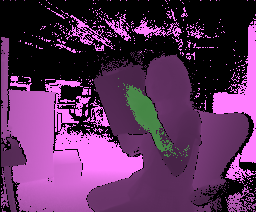}}
\end{minipage}
\begin{minipage}{0.18\linewidth}
\centerline{\includegraphics[scale=0.26]{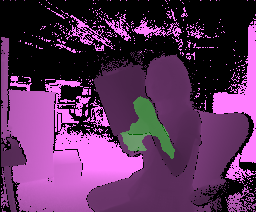}}
\end{minipage}
\begin{minipage}{0.18\linewidth}
\centerline{\includegraphics[scale=0.26]{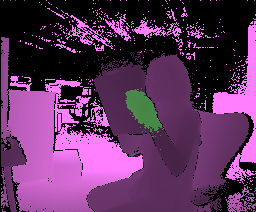}}
\end{minipage}
\begin{minipage}{0.18\linewidth}
\centerline{\includegraphics[scale=0.26]{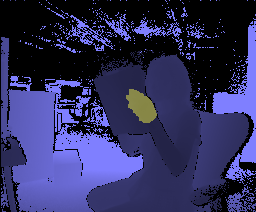}}
\end{minipage}
\\
\vspace{0.1cm}
\begin{minipage}{0.18\linewidth}
\centerline{\includegraphics[scale=0.26]{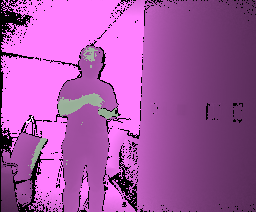}}
\end{minipage}
\begin{minipage}{0.18\linewidth}
\centerline{\includegraphics[scale=0.26]{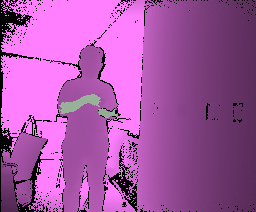}}
\end{minipage}
\begin{minipage}{0.18\linewidth}
\centerline{\includegraphics[scale=0.26]{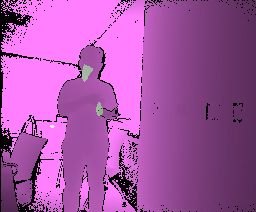}}
\end{minipage}
\begin{minipage}{0.18\linewidth}
\centerline{\includegraphics[scale=0.26]{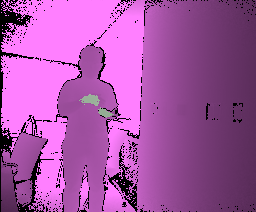}}
\end{minipage}
\begin{minipage}{0.18\linewidth}
\centerline{\includegraphics[scale=0.26]{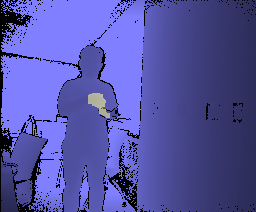}}
\end{minipage}
\\
\vspace{0.1cm}
\begin{minipage}{0.18\linewidth}
\centerline{\includegraphics[scale=0.26]{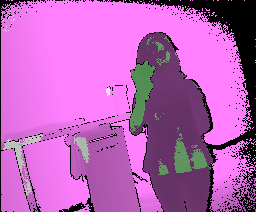}}
\centerline{\footnotesize (a) RF~\cite{shottoncvpr, tompson}} 
\end{minipage}
\begin{minipage}{0.18\linewidth}
\centerline{\includegraphics[scale=0.26]{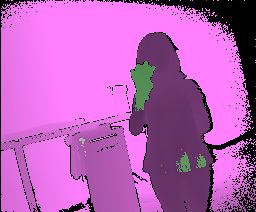}}
\centerline{\footnotesize (b) RF~\cite{shottoncvpr, tompson}+DB adjustment} 
\end{minipage}
\begin{minipage}{0.18\linewidth}
\centerline{\includegraphics[scale=0.26]{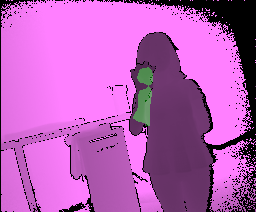}}
\centerline{\footnotesize (c) FCN-8s~\cite{longcvpr}    } 
\end{minipage}
\begin{minipage}{0.18\linewidth}
\centerline{\includegraphics[scale=0.26]{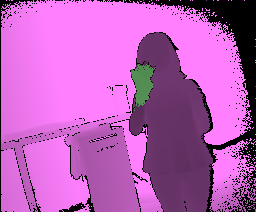}}
\centerline{\footnotesize (d) Proposed method} 
\end{minipage}
\begin{minipage}{0.18\linewidth}
\centerline{\includegraphics[scale=0.26]{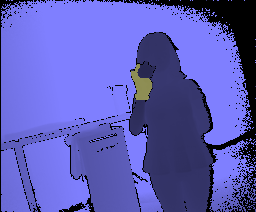}}
\centerline{\footnotesize (e) Ground truth label} 
\end{minipage}
   \caption{The qualitative comparison of the result for the HOI dataset. (a) Results of the random forest~\cite{shottoncvpr, tompson}. (b) Results of the random forest~\cite{shottoncvpr, tompson} with the decision boundary adjustment in~\ref{subsec:inference}. (c) Results of the FCN-8s model~\cite{longcvpr}. (d) Results of the proposed method. (e) Ground truth labels. The results and the ground truth labels are visualized on the depth maps with different color channels for better visualization.}
\label{fig:result}
\end{center}\end{figure*}

%% file: tbl_analysis_depth_hand.tex
\begin{table}[!t]
\caption{Analysis of the maximum depth in the proposed random forest using the HOI dataset.}
\label{tab:maximumDepthHOI}
\centering
\renewcommand{\arraystretch}{1.1}
\begin{tabu} to 0.5\textwidth {c|c|c|c|c|c|c}
\hline
Maximum & \# of  & \multicolumn{3}{c|}{Score} & Time & GPU \\
\cline{3-5}
depth 	 & nodes & Precision & Recall & $F_1$ score & ($ms$) & ($MB$)\\
\hline\hline
1		& 3 & 10.9 & 83.5 & 19.4 & 0.67 & 70  \\
\hline
5		& 46 & 36.4 & 54.2 & 43.6 & 0.77 & 71  \\
\hline
10 		& 1157 & 43.5 & 66.5 & 52.6 & 0.96 & 71  \\
\hline
15		& 15440 & 50.2 & 73.2 & 59.6 & 1.12 & 73   \\
\hline 
20		& 57176 & 56.9 & 74.0 & 64.4 & 1.22 & 81 \\
\hline 
25		& 100523 & 59.8 & 73.8 & 66.1 & 1.26 & 90  \\
\hline 
\end{tabu}
\end{table}


%% file: tbl_result_cityscapes.tex
\begin{table*}[!t]
\caption{The quantitative results of the Cityscapes dataset.}
\label{tab:cityscapesResult}
\centering
\renewcommand{\arraystretch}{1.1}
\begin{tabu} to 0.96\textwidth { c|X[c,m]|X[c,m]|X[c,m]|X[c,m]|X[c,m]|X[c,m]|X[c,m]}
\hline
\multicolumn{3}{c|}{Method} & Input & \multicolumn{2}{c|}{Accuracy (IU)} & Time & GPU memory \\
\cline{1-3}
\cline{5-6}
Method & \# of trees & Bilateral filter & resolution & Category & Class & ($ms$) &  ($MB$)\\
\hline\hline
OCNet~\cite{OCNet}						& -	& -	& \multirow{3}{*}{Multiscale} & 91.6 & 81.7 & 58870 & 21994 \\
RefineNet~\cite{refinenet}					& -	& -	&  & 87.9 & 73.6 & 22950 & 17647 \\
ICNet~\cite{icnet}						& -	& -	&  & - & 70.6 & 212 & 1549 \\
\hline
CCNet~\cite{CCNet}		& -	& -	& \multirow{7}{*}{2048$\times$1024} & - & 81.4 & 9233 & 12016 \\
PSPNet~\cite{pspnet}						& -	& -	&  & 90.6 & 80.2 & 15254 & 63288 \\
FCN-8s~\cite{longcvpr, longpami}			& -	& -	& & 85.7 & 65.3 & 1365 & 5800 \\
CRF as RNN~\cite{Zheng}					& -	& -	& & 82.7 & 62.5 & 10704 & 31836 \\
ENet~\cite{enet}							& -	& -	& & 80.4 & 58.3 & 2966 & 284 \\
SegNet basic~\cite{segnet}				& -	& -	& & 79.1 & 57.0 & 1392 & 4564 \\
SegNet extended~\cite{segnet}				& -	& -	& & 79.8 & 56.1 & 1992 & 9158
 \\
\hline 
Proposed									& 5 & 23$\times$23 & 512$\times$256 & 60.2 & 30.0 & 30 & 1271 \\
\hline
\end{tabu}
\end{table*}

%% file: fig_result_cityscapes.tex
\begin{figure*}[!t] \begin{center}
\begin{minipage}{0.19\linewidth}
\centerline{\includegraphics[scale=0.045]{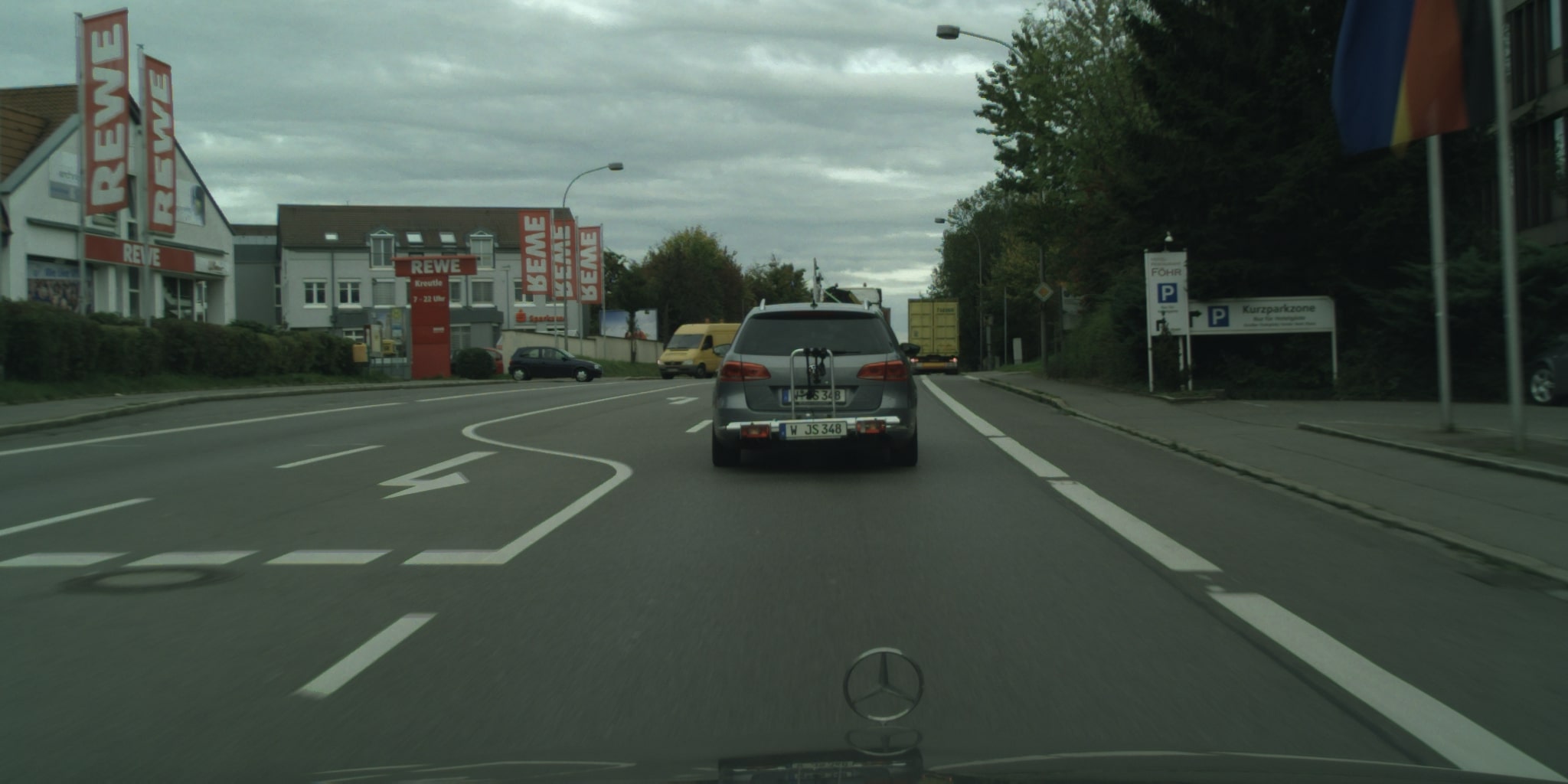}}
\end{minipage}
\begin{minipage}{0.19\linewidth}
\centerline{\includegraphics[scale=0.045]{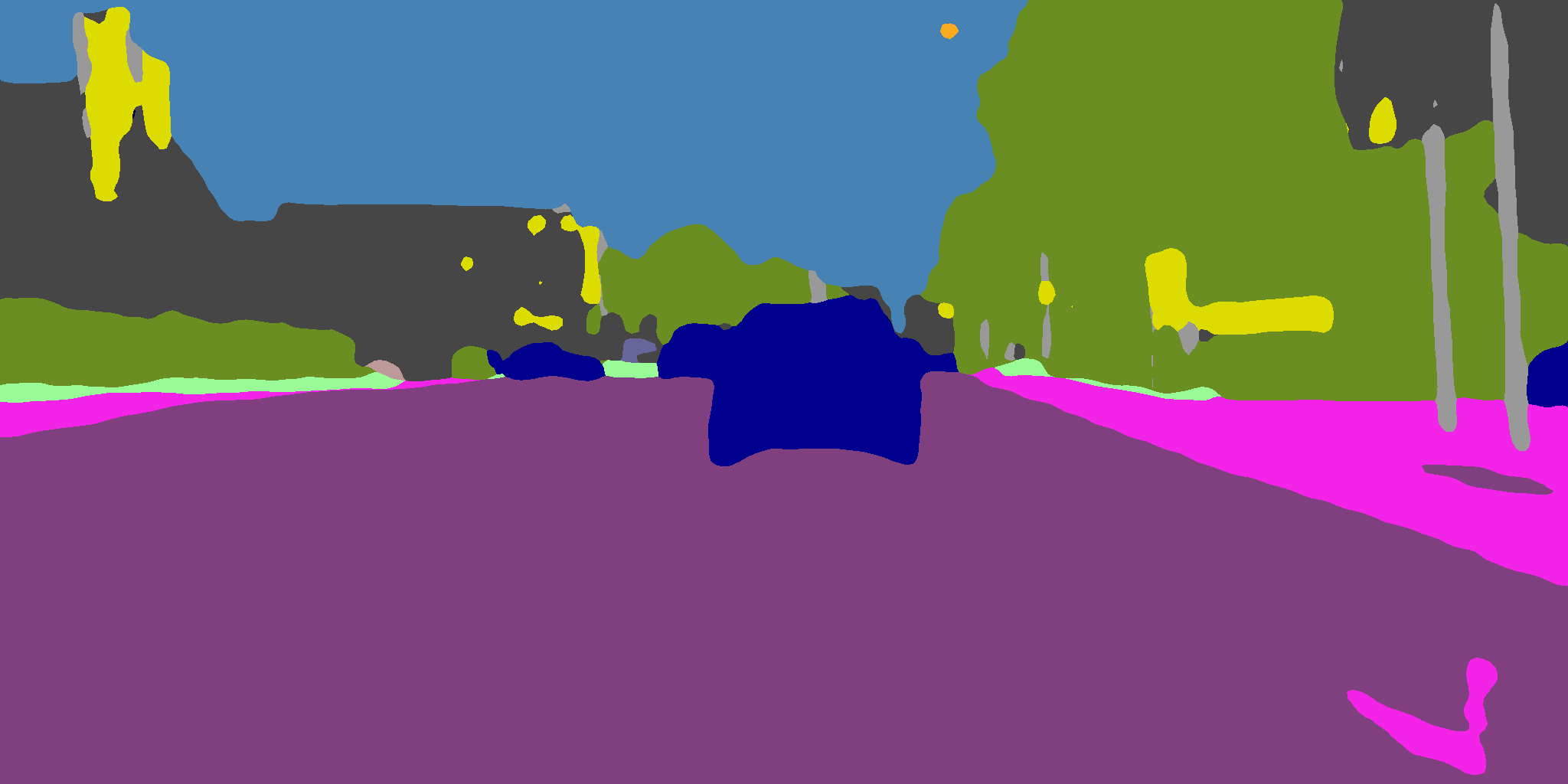}}
\end{minipage}
\begin{minipage}{0.19\linewidth}
\centerline{\includegraphics[scale=0.045]{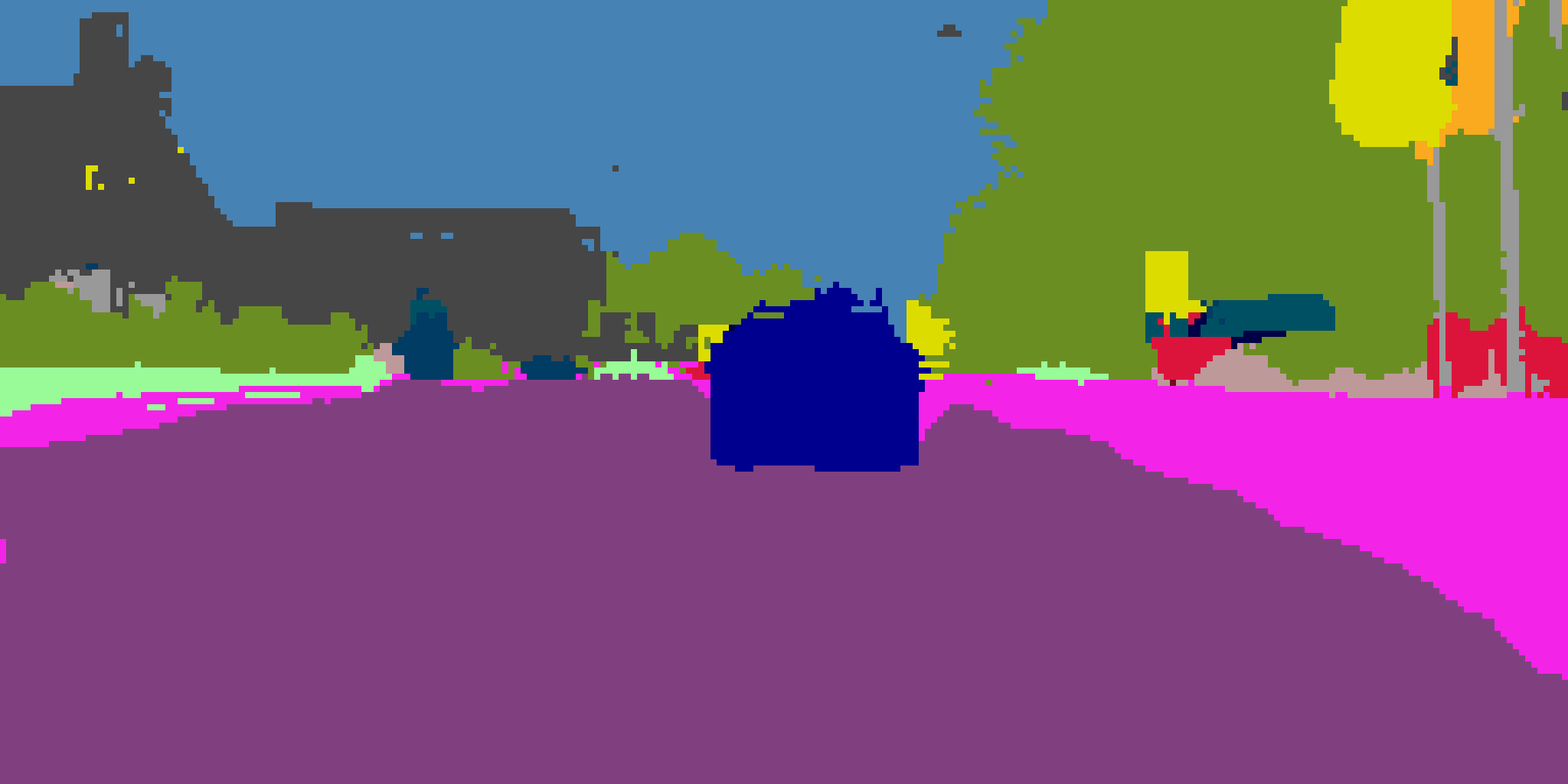}}
\end{minipage}
\begin{minipage}{0.19\linewidth}
\centerline{\includegraphics[scale=0.045]{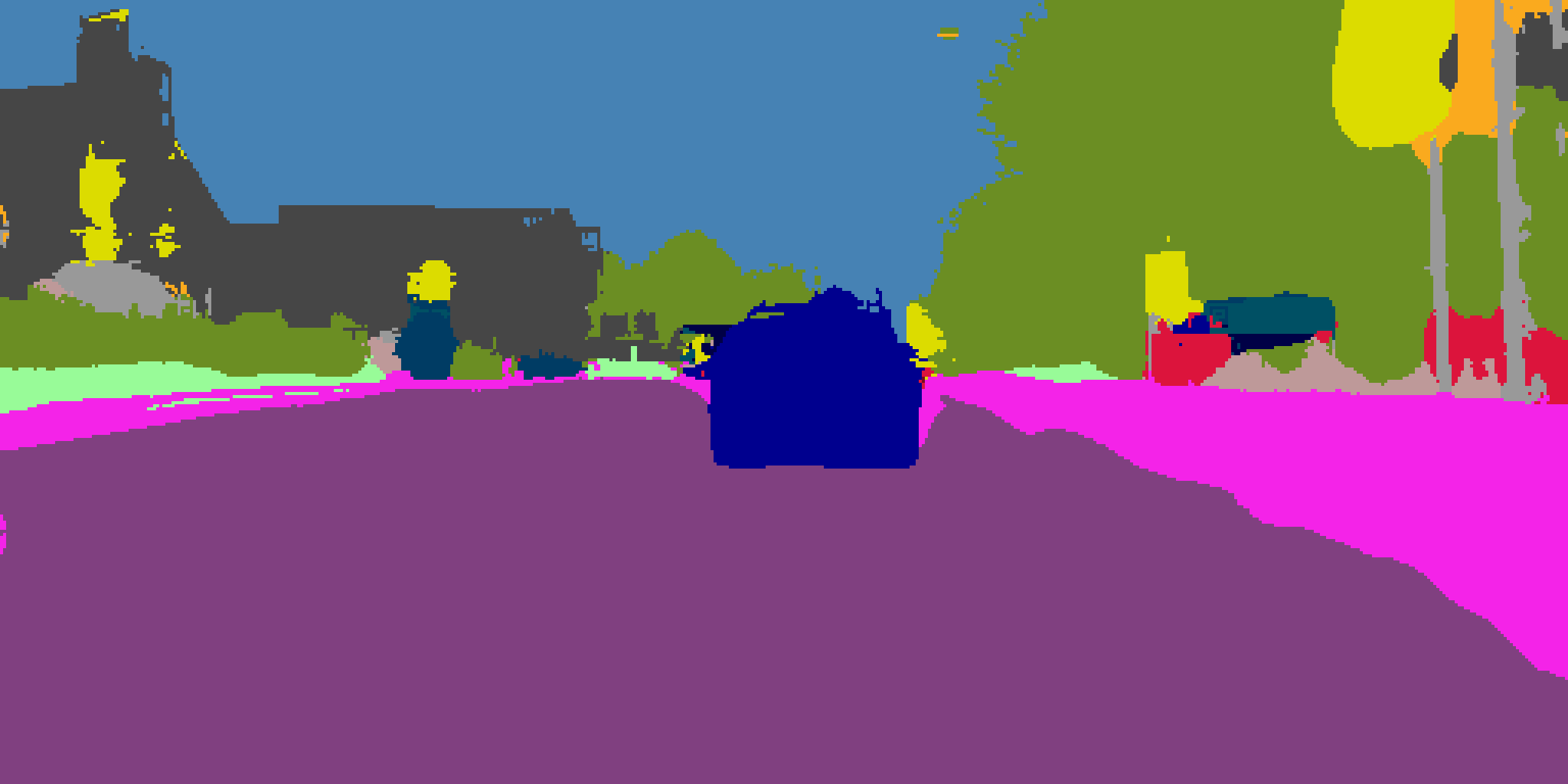}}
\end{minipage}
\begin{minipage}{0.19\linewidth}
\centerline{\includegraphics[scale=0.045]{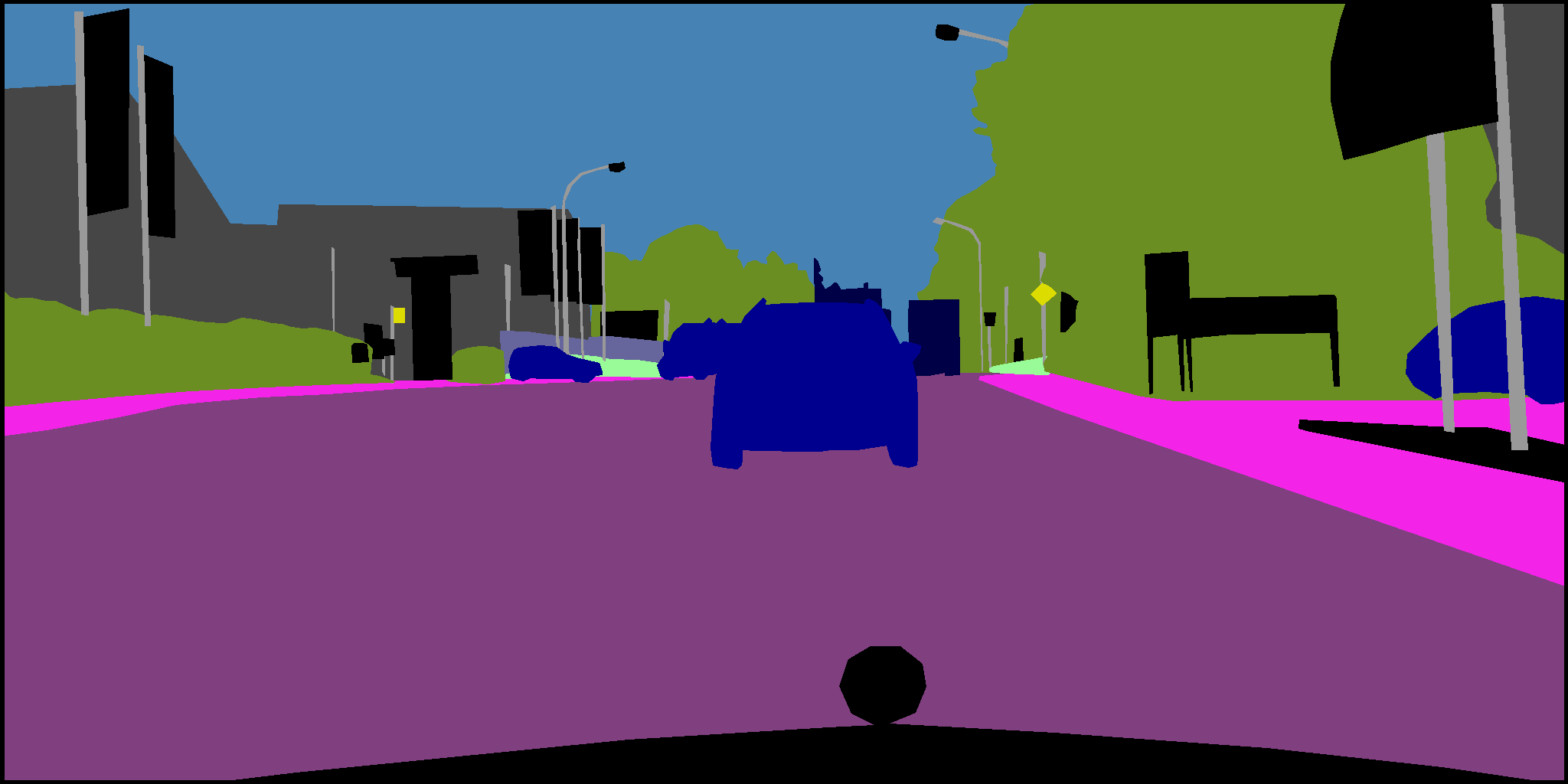}}
\end{minipage}
\\
\vspace{0.1cm}
\begin{minipage}{0.19\linewidth}
\centerline{\includegraphics[scale=0.045]{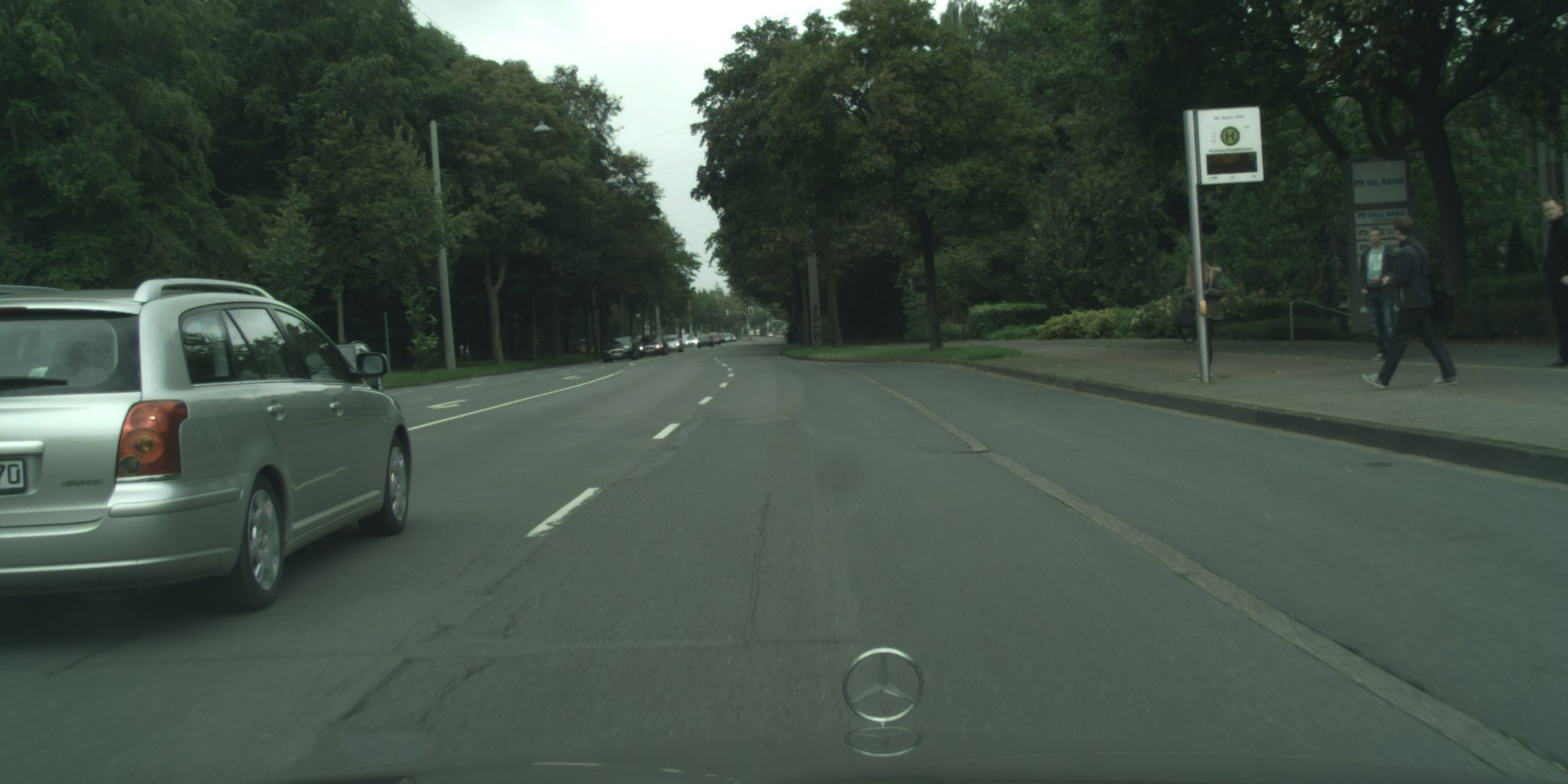}}
\end{minipage}
\begin{minipage}{0.19\linewidth}
\centerline{\includegraphics[scale=0.045]{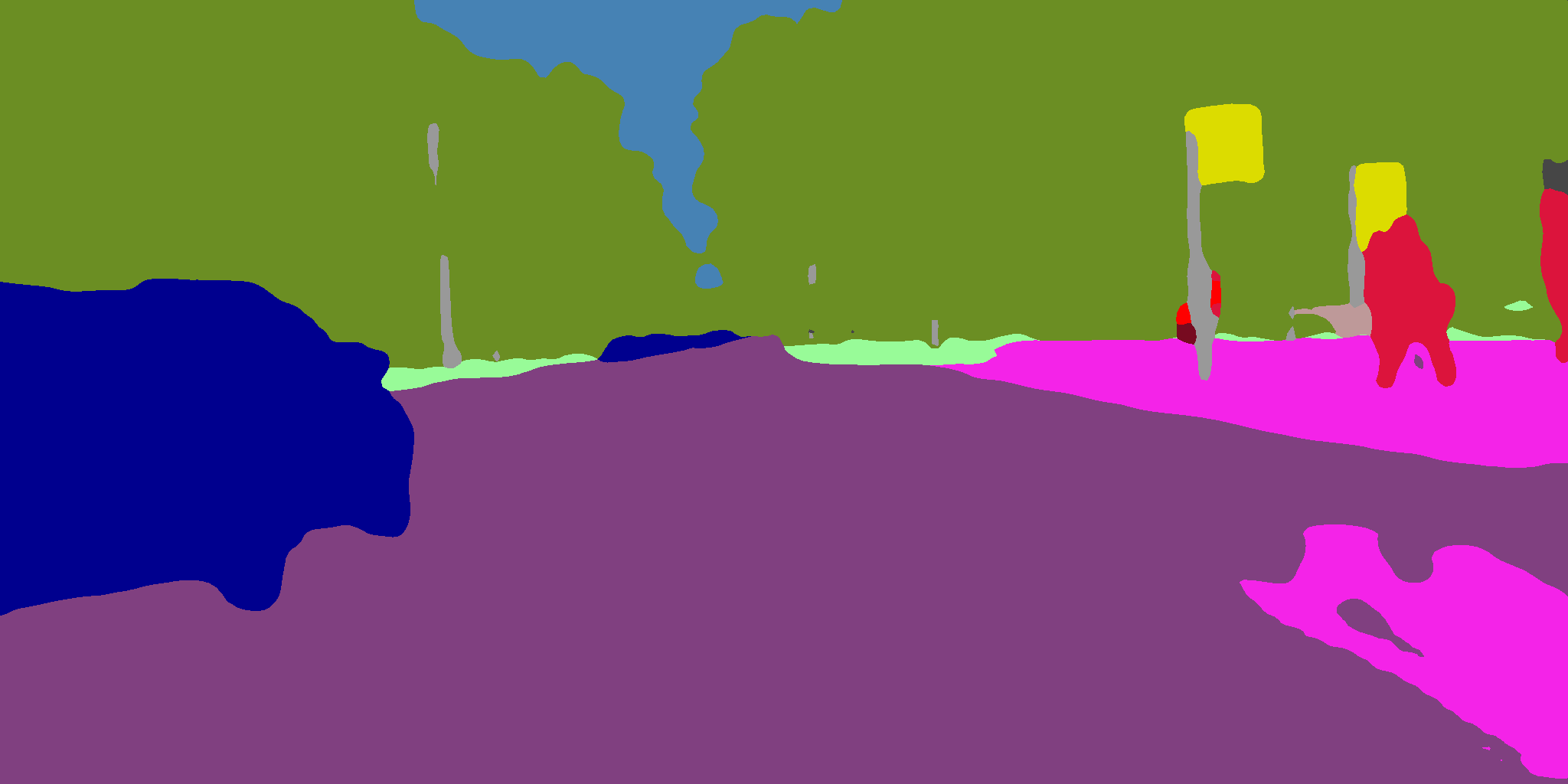}}
\end{minipage}
\begin{minipage}{0.19\linewidth}
\centerline{\includegraphics[scale=0.045]{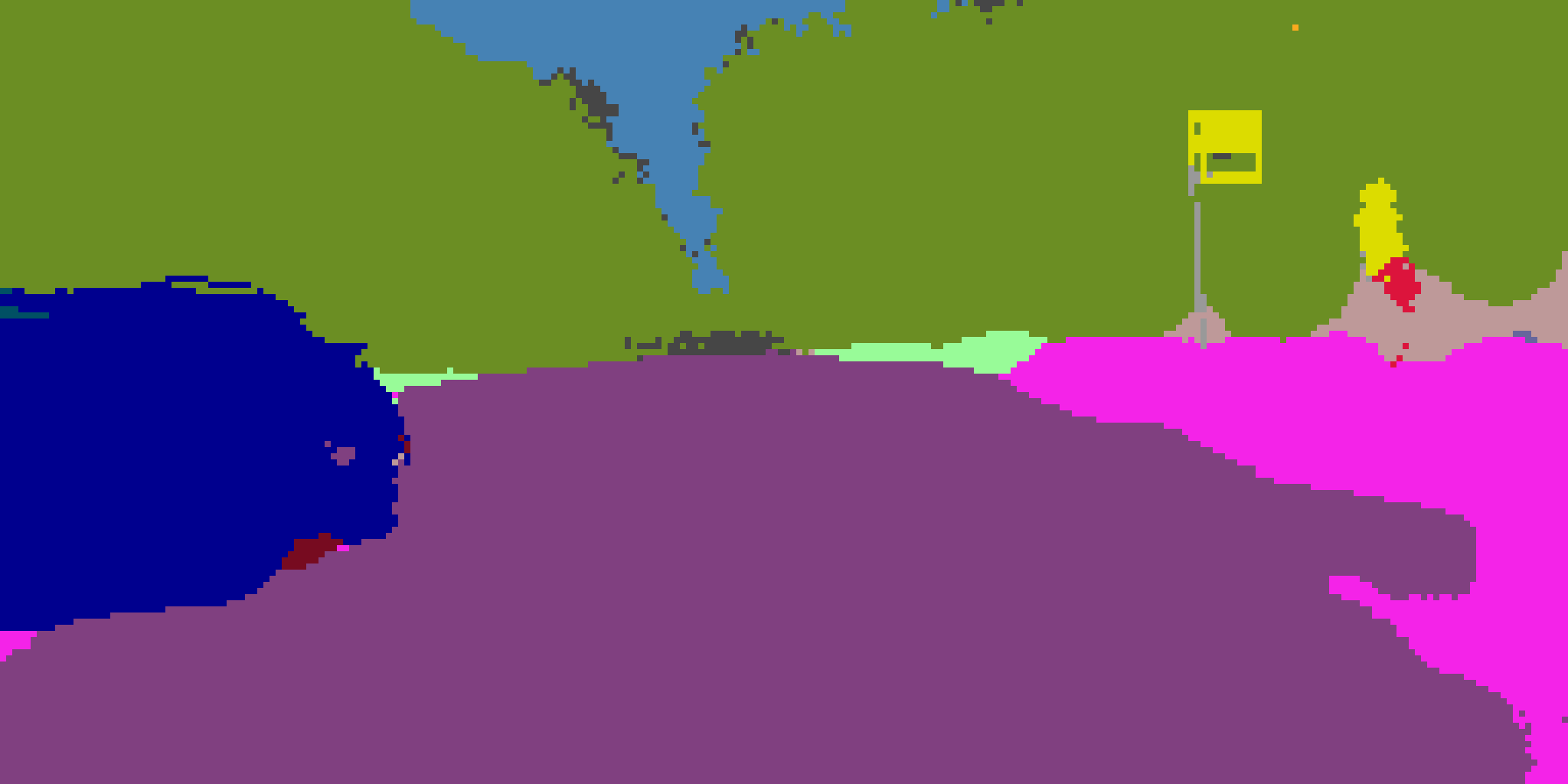}}
\end{minipage}
\begin{minipage}{0.19\linewidth}
\centerline{\includegraphics[scale=0.045]{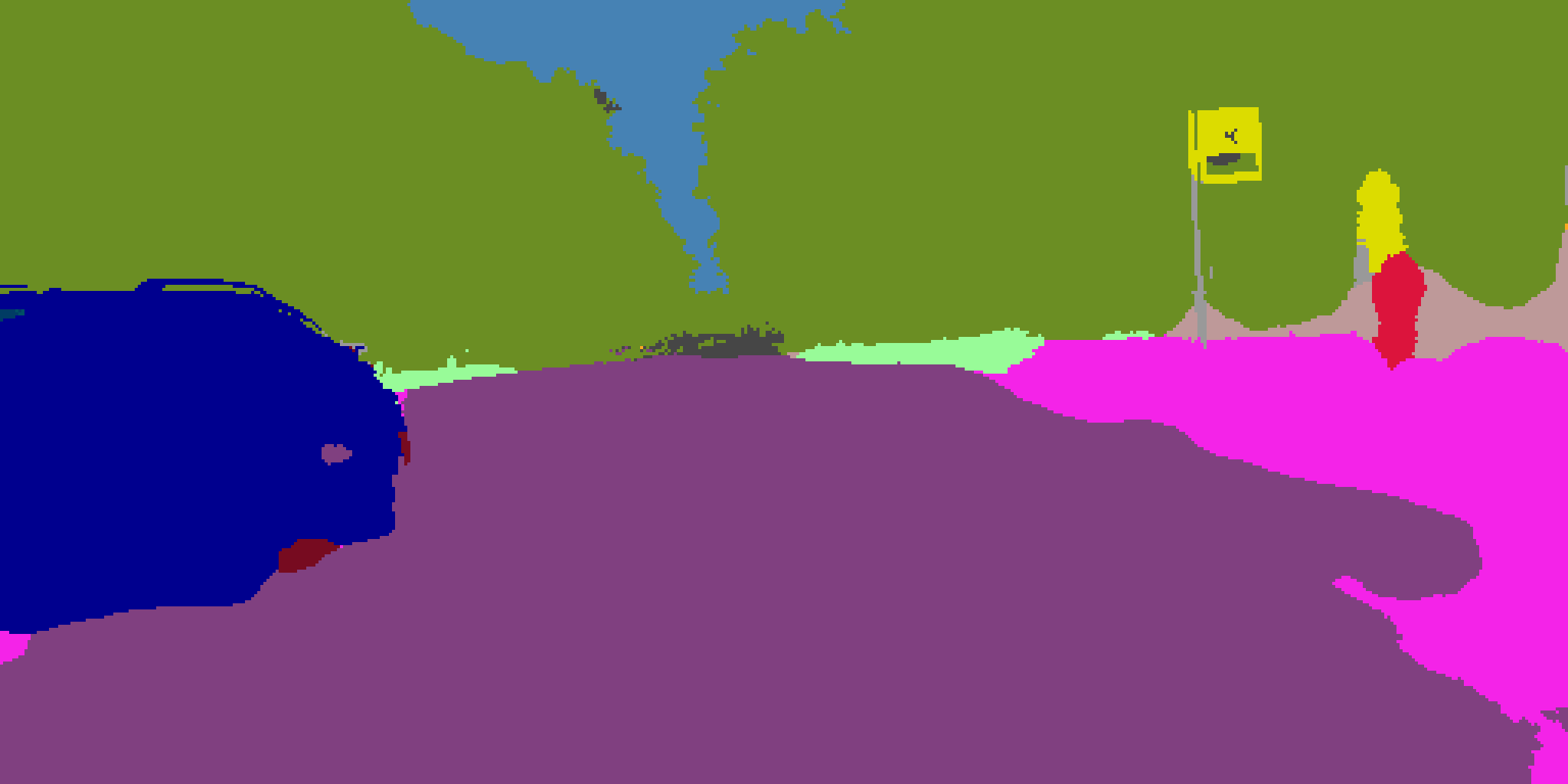}}
\end{minipage}
\begin{minipage}{0.19\linewidth}
\centerline{\includegraphics[scale=0.045]{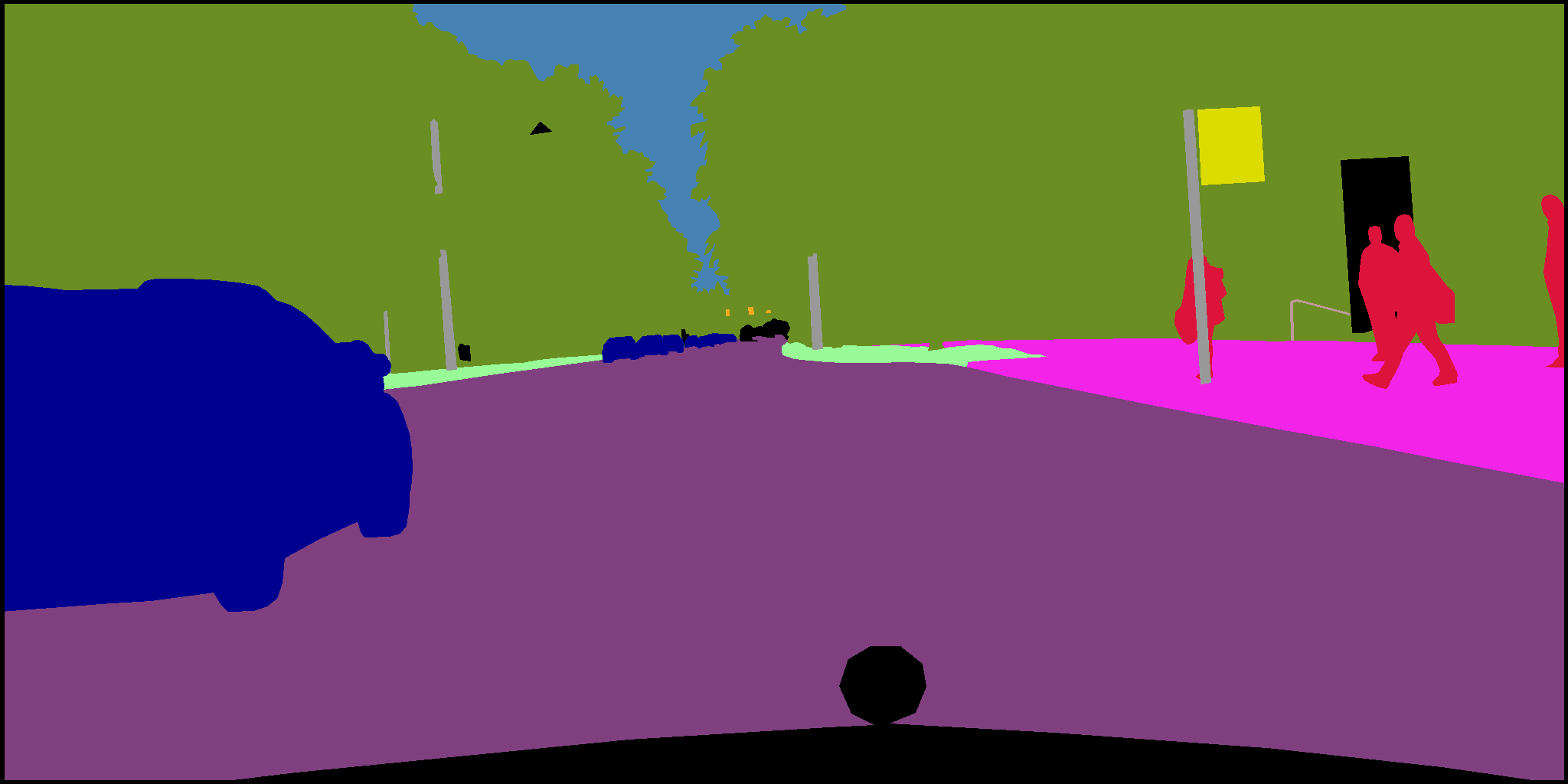}}
\end{minipage}
\\
\vspace{0.1cm}
\begin{minipage}{0.19\linewidth}
\centerline{\includegraphics[scale=0.045]{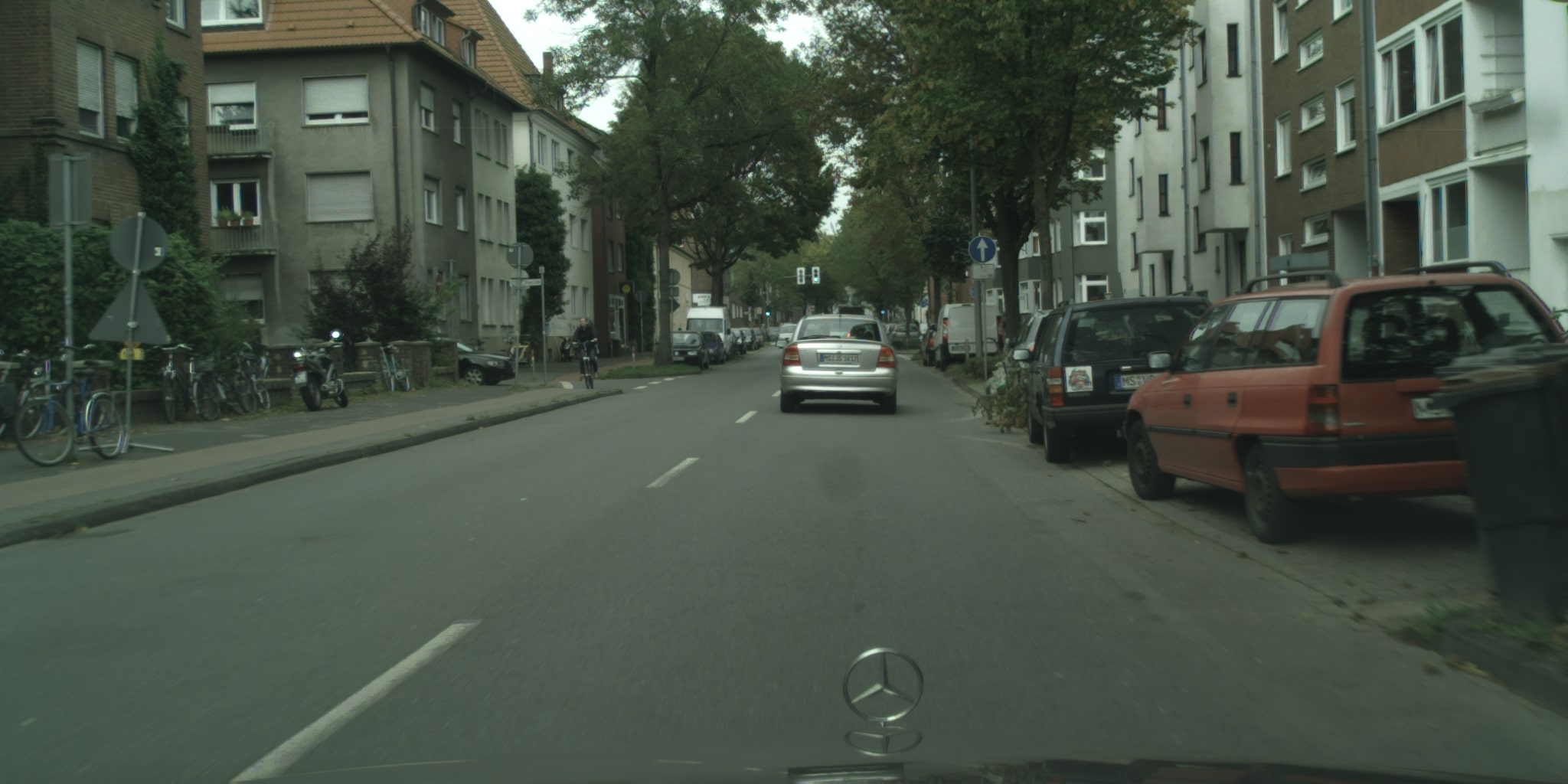}}
\end{minipage}
\begin{minipage}{0.19\linewidth}
\centerline{\includegraphics[scale=0.045]{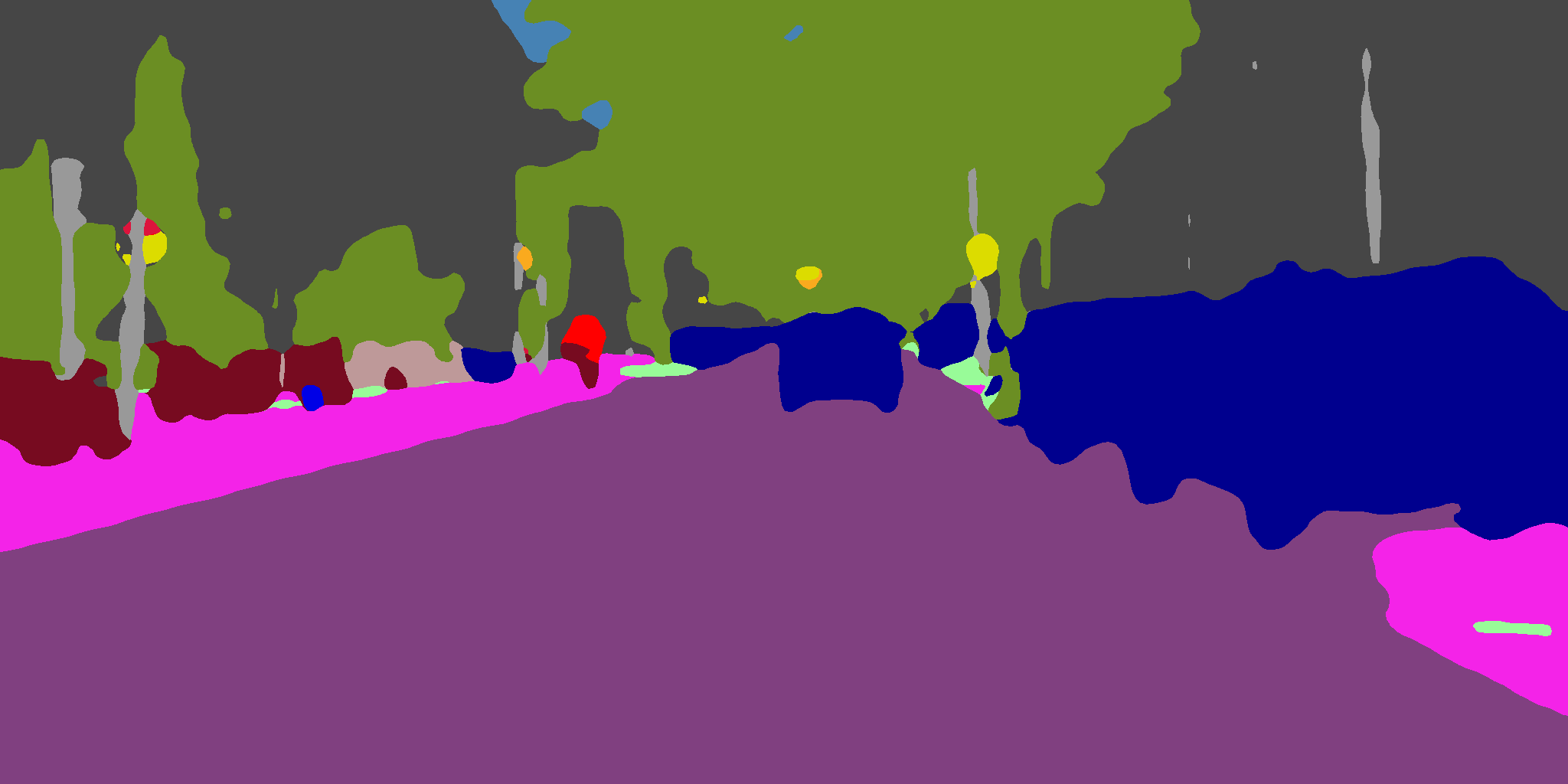}}
\end{minipage}
\begin{minipage}{0.19\linewidth}
\centerline{\includegraphics[scale=0.045]{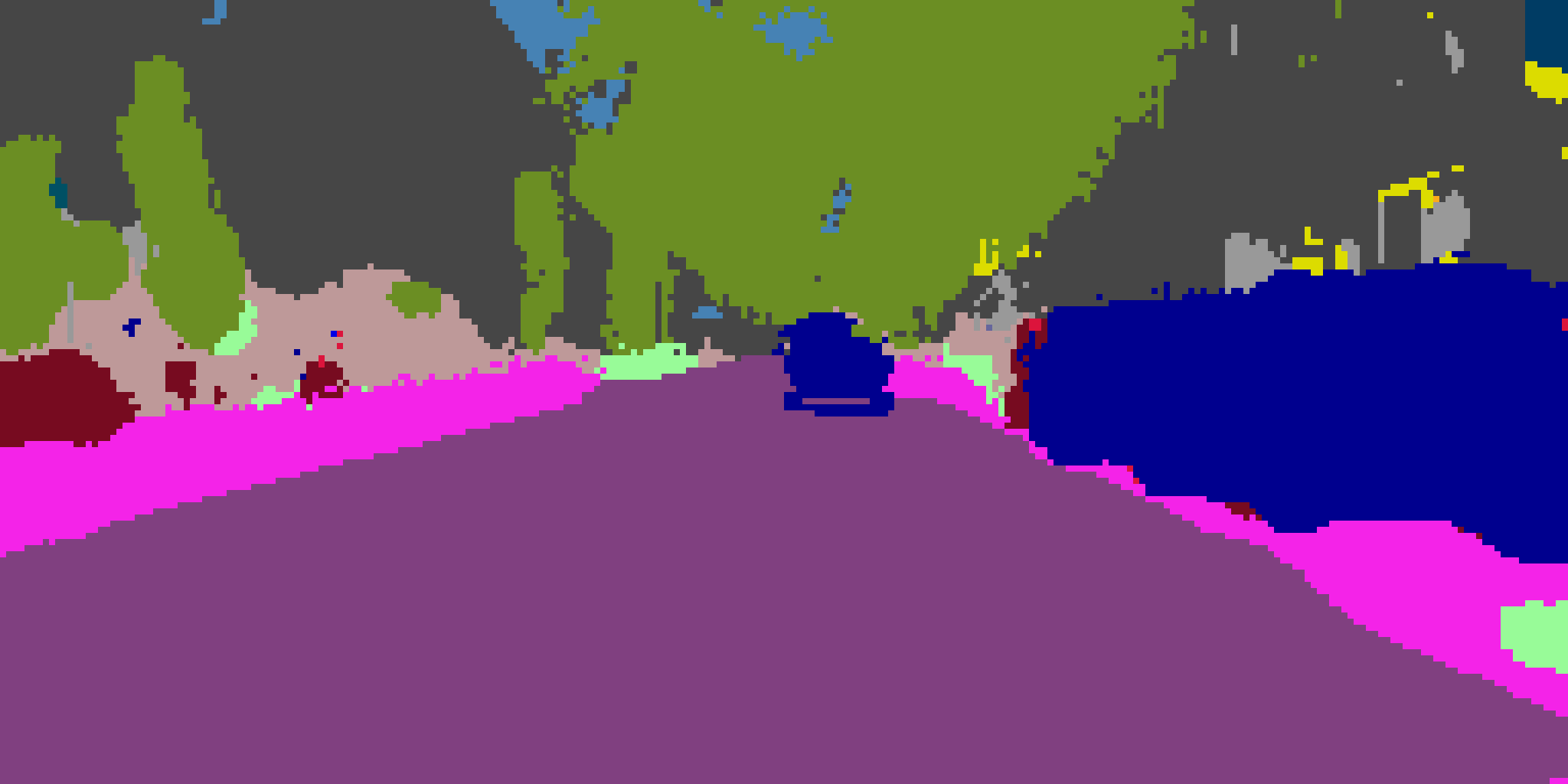}}
\end{minipage}
\begin{minipage}{0.19\linewidth}
\centerline{\includegraphics[scale=0.045]{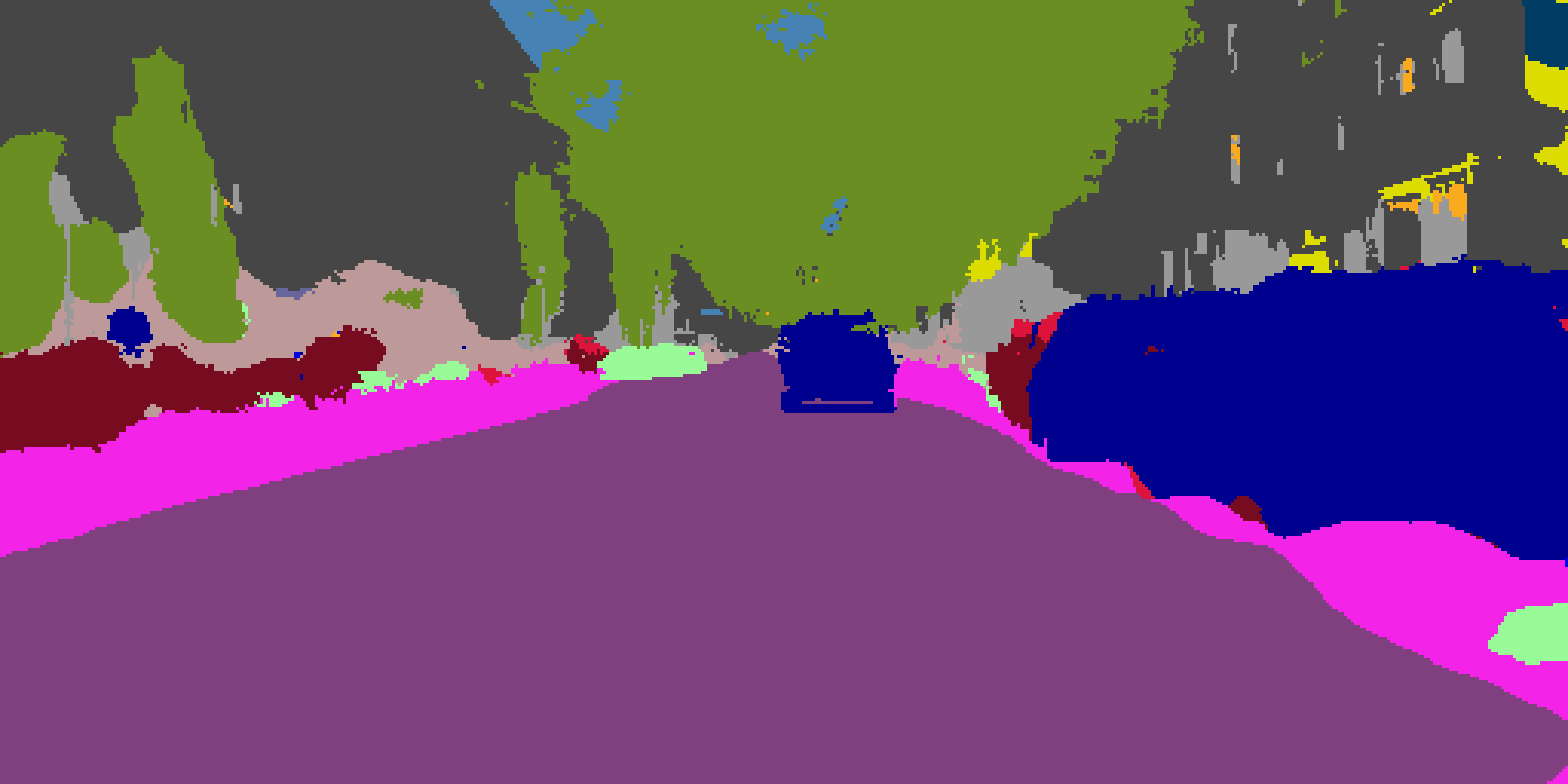}}
\end{minipage}
\begin{minipage}{0.19\linewidth}
\centerline{\includegraphics[scale=0.045]{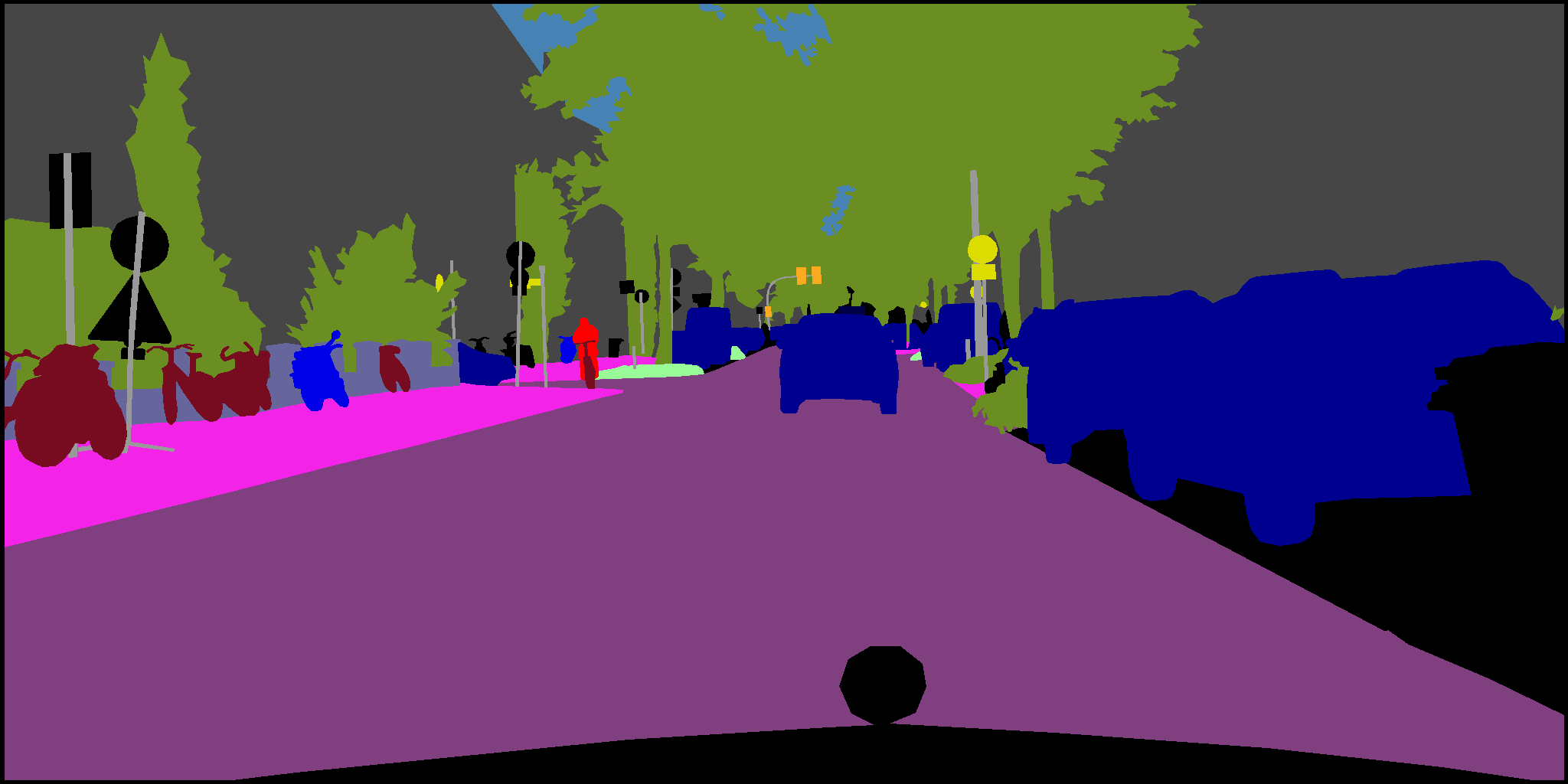}}
\end{minipage}
\\
\vspace{0.1cm}
\begin{minipage}{0.19\linewidth}
\centerline{\includegraphics[scale=0.045]{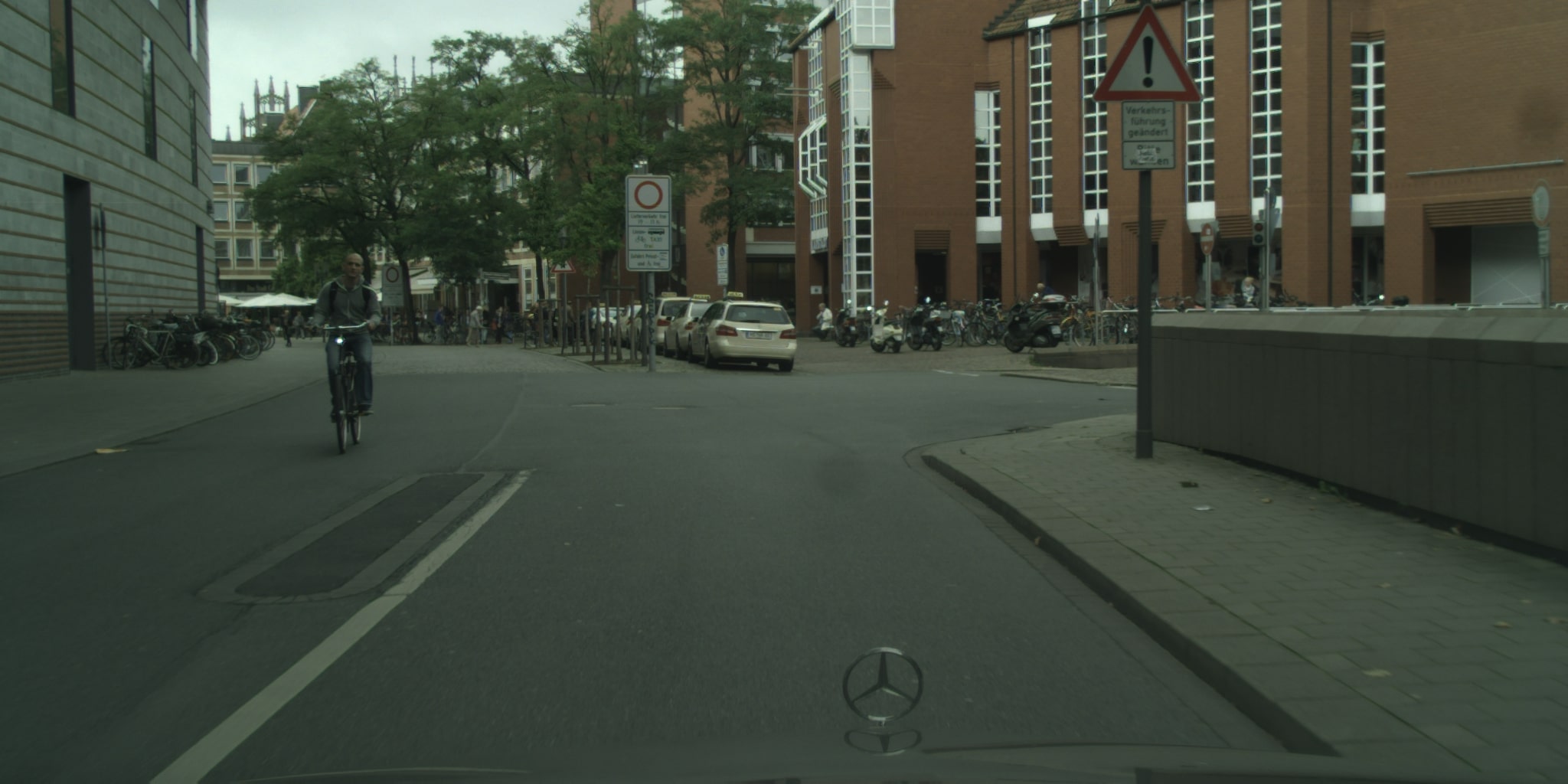}}
\centerline{\footnotesize (a) Input image}
\end{minipage}
\begin{minipage}{0.19\linewidth}
\centerline{\includegraphics[scale=0.045]{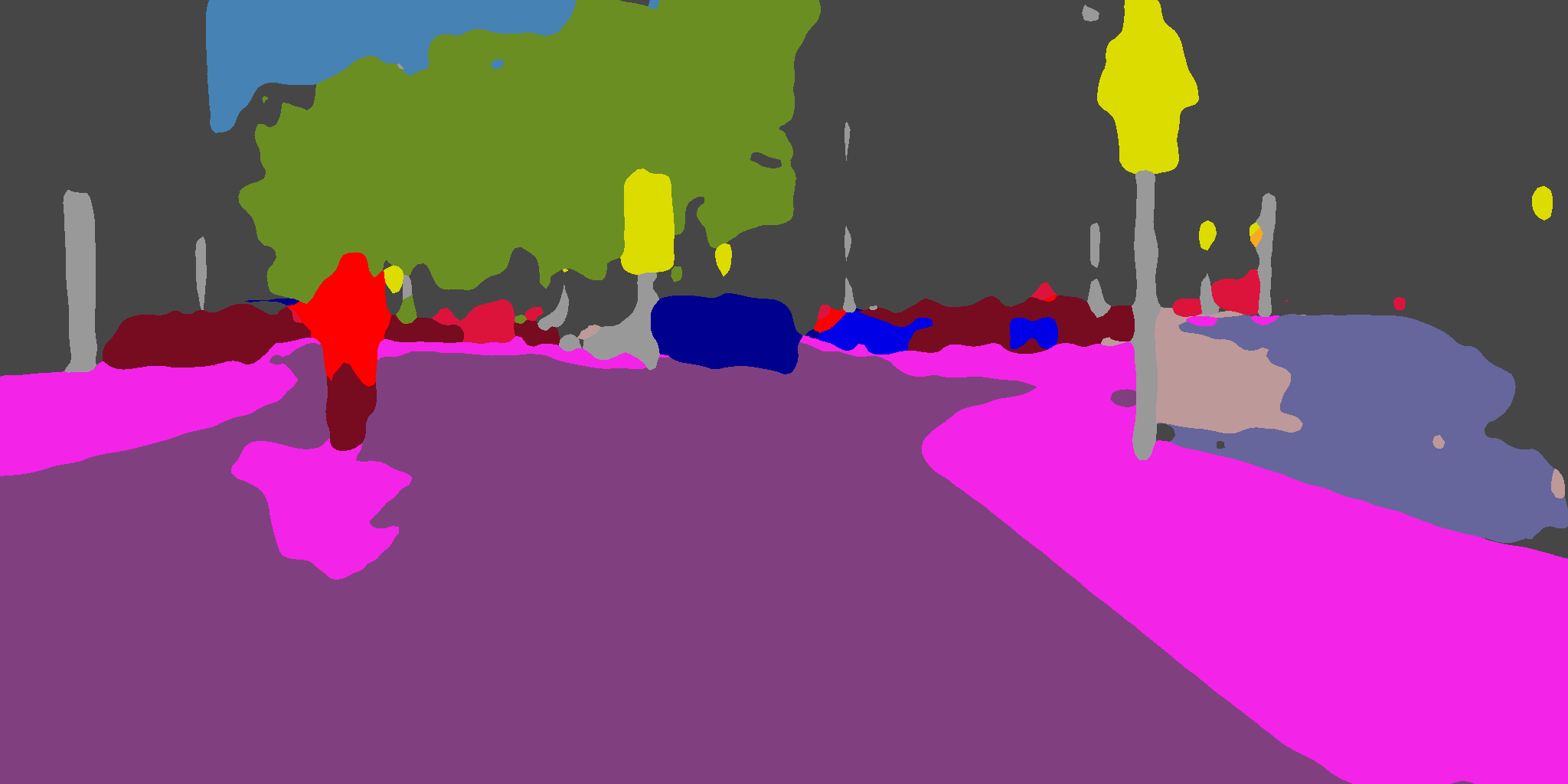}}
\centerline{\footnotesize (b) FCN-8s~\cite{longcvpr, longpami}} 
\end{minipage}
\begin{minipage}{0.19\linewidth}
\centerline{\includegraphics[scale=0.045]{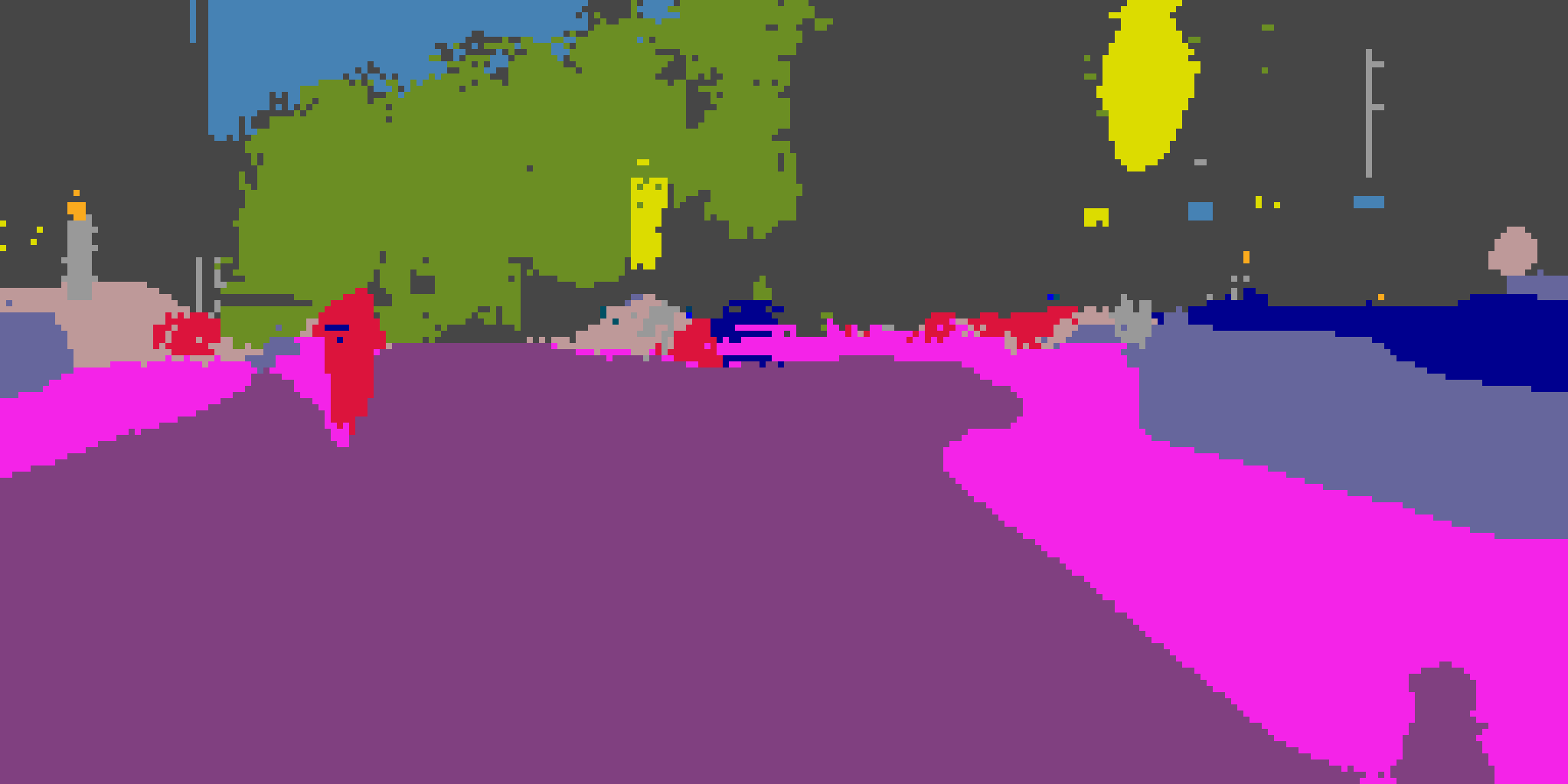}}
\centerline{\footnotesize (c) Proposed method} 
\end{minipage}
\begin{minipage}{0.19\linewidth}
\centerline{\includegraphics[scale=0.045]{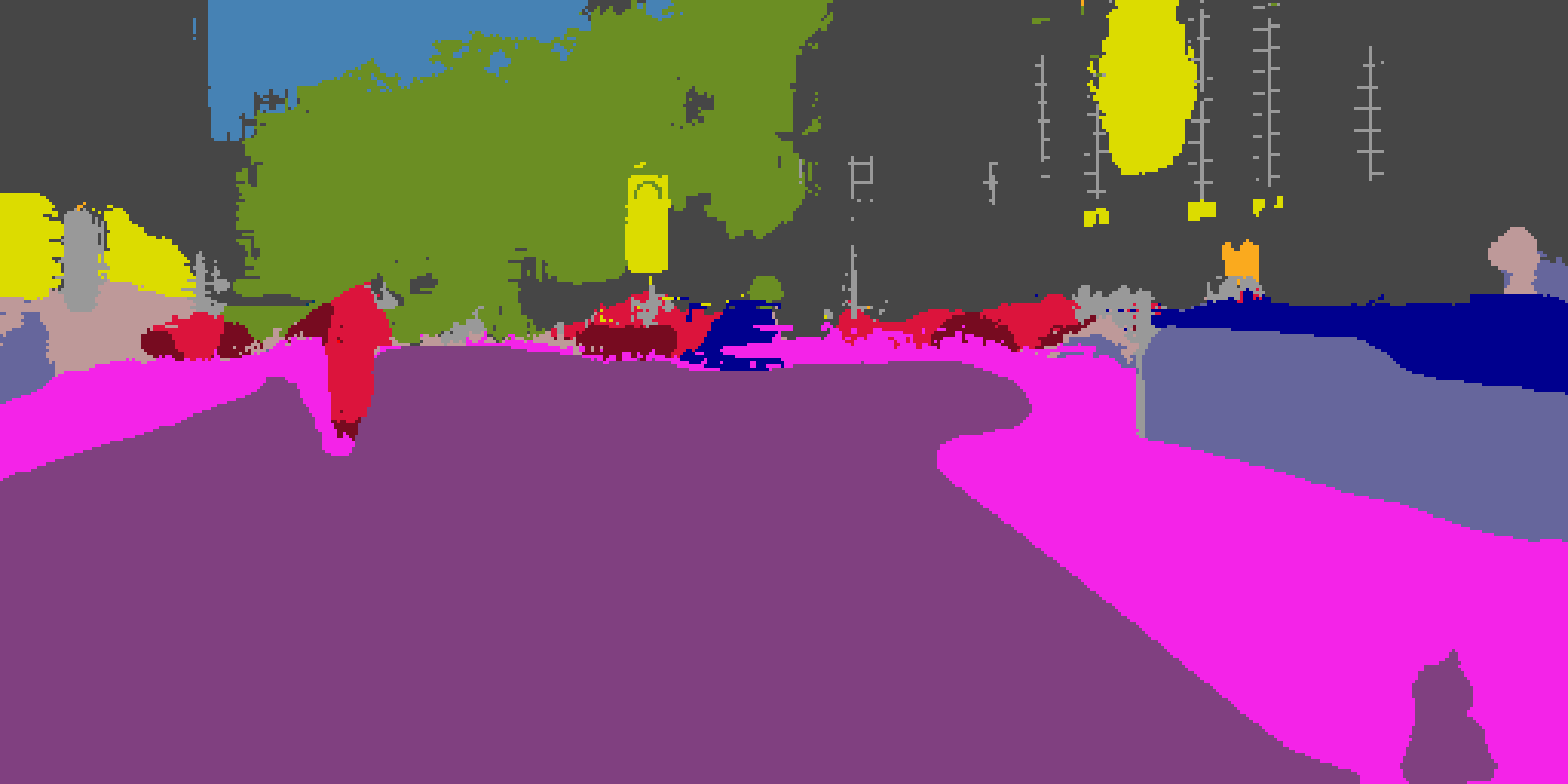}}
\centerline{\footnotesize (d) Proposed method} 
\end{minipage}
\begin{minipage}{0.19\linewidth}
\centerline{\includegraphics[scale=0.045]{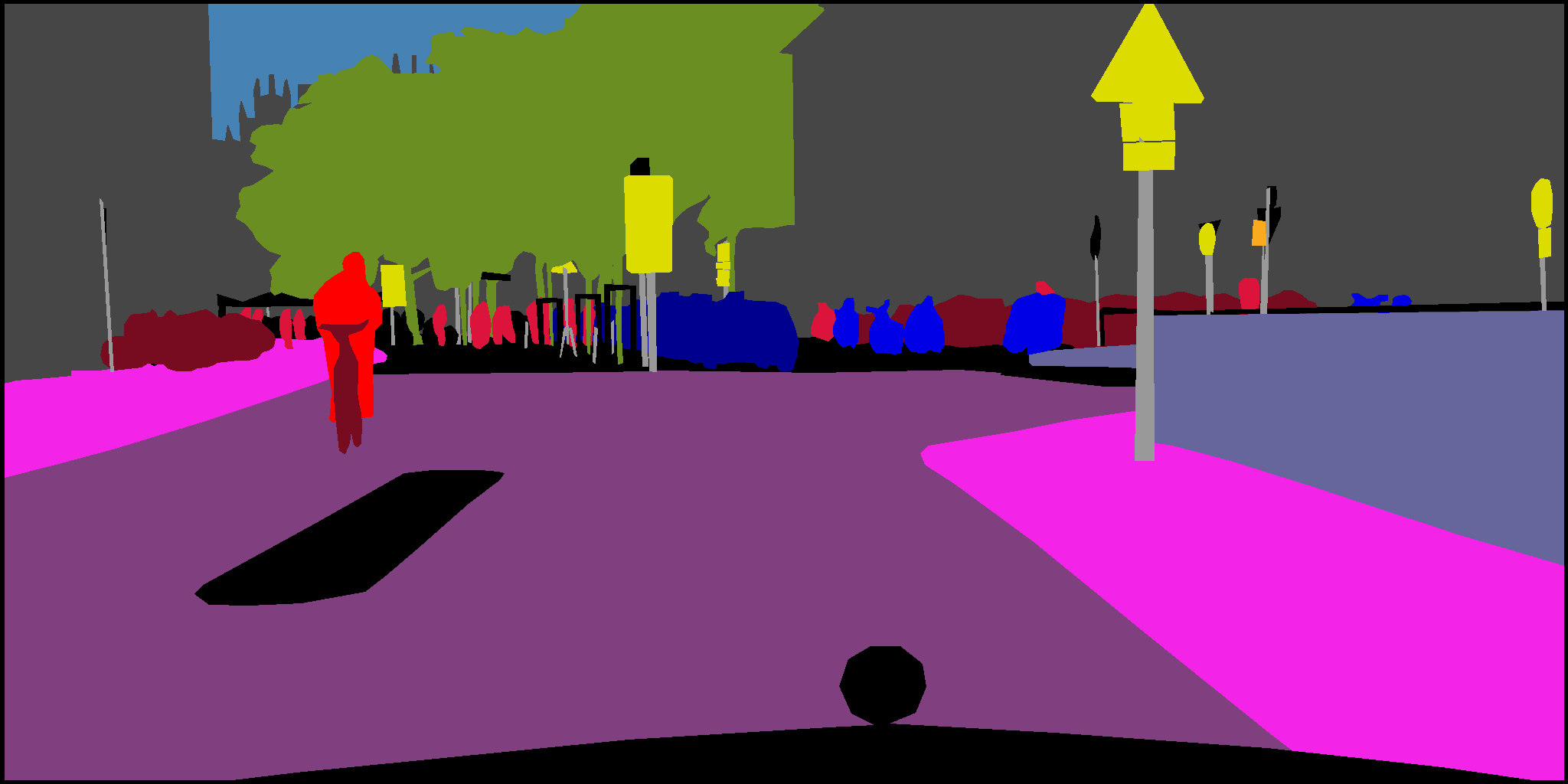}}
\centerline{\footnotesize (e) Ground truth label} 
\end{minipage}
   \caption{The qualitative comparison of the result for the Cityscapes dataset~\cite{cityscapes}. (a) Input image. (b) Results of the FCN-8s model~\cite{longcvpr, longpami}. (c) Results of the proposed method using five trees, bilateral filter of 13$\times$13, and the input resolution of 256$\times$128. (d) Results of the proposed method using five trees, bilateral filter of 23$\times$23, and the input resolution of 512$\times$256. (e) Ground truth labels.}
\label{fig:cityscapesResult}
\end{center}\end{figure*}

%% file: tbl_analysis_cityscapes.tex
\begin{table*}[!t]
\caption{Analysis of the proposed method using the validation set of the Cityscapes dataset.}
\label{tab:cityscapesAnalysis}
\centering
\renewcommand{\arraystretch}{1.1}
\begin{tabu} to 0.96\textwidth { c|X[c,m]|X[c,m]|X[c,m]|X[c,m]|X[c,m]|X[c,m]|X[c,m]}
\hline
\multicolumn{3}{c|}{Method} & Input & \multicolumn{2}{c|}{Accuracy (IU)} & Time & GPU memory \\
\cline{1-3}
\cline{5-6}
Method & \# of trees & Bilateral filter & resolution & Category & Class & ($ms$) &  ($MB$)\\
\hline\hline
\multirow{20}{*}{Proposed}
									& \multirow{3}{*}{1} & -	& \multirow{6}{*}{2048$\times$1024} & 40.0 & 16.9 & \multirow{3}{*}{75} & \multirow{3}{*}{329} \\
 									& & 15$\times$15 & & 50.0 & 23.0 &  &  \\
 									& & 91$\times$91 & & 53.3 & 25.5 &  &  \\
\cline{2-3}\cline{5-8}
									& \multirow{3}{*}{5} & -	&  & 45.9 & 20.5 & \multirow{3}{*}{385} & \multirow{3}{*}{1319} \\
									& & 13$\times$13 & & 52.8 & 25.1 &  &  \\
									& & 91$\times$91 & & 54.6 & 26.4 &  &  \\
\cline{2-8}
									& \multirow{3}{*}{1} & -	& \multirow{6}{*}{1024$\times$512} & 41.1 & 17.4 & \multirow{3}{*}{21} & \multirow{3}{*}{290} \\
									& & 15$\times$15 & & 52.0 & 24.4 &  &  \\
 									& & 55$\times$55 & & 53.6 & 25.7 &  &  \\
\cline{2-3}\cline{5-8}

									& \multirow{3}{*}{5} & -	&  & 46.8 & 21.0 &\multirow{3}{*}{106} & \multirow{3}{*}{1280} \\
									& & 13$\times$13 & & 54.1 & 26.0 &  &  \\
									& & 43$\times$43 & & 55.0 & 26.7 &  &  \\
\cline{2-8}
									& \multirow{3}{*}{1} & -	& \multirow{6}{*}{512$\times$256} & 41.8 & 17.8 & \multirow{3}{*}{6} & \multirow{3}{*}{281} \\
									& & 15$\times$15 & & 53.3 & 25.4 &  &  \\
 									& & 27$\times$27 & & 53.7 & 25.8 &  &  \\
\cline{2-3}\cline{5-8}

									& \multirow{3}{*}{5} & -	&  & 47.4 & 21.4 &\multirow{3}{*}{30} & \multirow{3}{*}{1271} \\
									& & 13$\times$13 & & 54.9 & 26.6 &  &  \\
									& & 23$\times$23 & & 55.1 & 26.8 &  &  \\
\cline{2-8}
									& 1 & 15$\times$15 & \multirow{2}{*}{256$\times$128} & 52.8 & 25.2 & 2 & 279 \\
\cline{2-3}\cline{5-8}
									& 5 & 13$\times$13 & & 54.2 & 26.3 & 8 & 1269  \\
\hline
\end{tabu}
\end{table*}

%% file: tbl_analysis_PSO.tex
\begin{table*}[!t]
\caption{Analysis on learning features using the proposed unconstrained representation and the particle swarm optimization.}
\label{tab:pso}
\centering
\renewcommand{\arraystretch}{1.1}
\begin{tabu} to 0.99\textwidth {X[c,m]|X[c,m]|X[c,m]|X[c,m]|X[c,m]|X[c,m]|X[c,m]|X[c,m]}
\hline
\multicolumn{2}{c|}{Method} & \multirow{2}{*}{\# of nodes} & \multirow{2}{*}{Bilateral filter} & \multicolumn{2}{c|}{Accuracy (IU)} & Time & GPU memory \\
\cline{1-2}\cline{5-6}
Representation & Optimization & & & Category & Class &  ($ms$) & ($MB$)\\
\hline \hline
&	& \multirow{4}{*}{1,068,413} & - & 29.1 & 10.4 & \multirow{4}{*}{149} & \multirow{4}{*}{561} \\
Convolution 	 	 & Random	& & 11$\times$11 & 36.2 & 14.6 & & \\
filter (3$\times$3) & search	& & 51$\times$51 & 38.5 & 16.7 & & \\
				 & 			& & 71$\times$71 & 38.7 & 16.9 & & \\
\hline
&	& \multirow{4}{*}{855,974} & - & 31.5 & 11.3 & \multirow{4}{*}{102} & \multirow{4}{*}{398} \\
Proposed 		 & Random	& & 11$\times$11 & 40.4 & 15.7 & & \\
unconstrained	 & search	& & 51$\times$51 & 44.9 & 18.6 & & \\
				 &			& & 71$\times$71 & 45.6 & 19.1 & & \\
\hline
& 	& \multirow{4}{*}{1,170,745} & - & 29.1 & 10.4 & \multirow{4}{*}{152} & \multirow{4}{*}{603} \\
Convolution		 & Proposed 	& & 11$\times$11 & 37.0 & 15.1 & & \\
filter (3$\times$3) & PSO		& & 51$\times$51 & 39.4 & 17.2 & & \\
				 &			& & 71$\times$71 & 39.5 & 17.4 & & \\
\hline
& 	& \multirow{4}{*}{767,953} & - & 37.2 & 14.9 & \multirow{4}{*}{83} & \multirow{4}{*}{371} \\
Proposed 		& Proposed	& & 11$\times$11 & 47.2 & 21.0  &  \\
unconstrained 	& PSO		& & 51$\times$51 &  52.7 & 24.6 &  \\
				&			& & 71$\times$71 &  53.3 & 25.0 &  \\
\hline
\end{tabu}
\end{table*}


%% file: tbl_analysis_resize.tex
\begin{table}[!t]
\caption{Analysis on applying neural networks to input images with different resolution.}
\label{tab:cityscapesResize}
\centering
\renewcommand{\arraystretch}{1.1}
\begin{tabu} to 0.5\textwidth { c|c|X[c,m]|X[c,m]|X[c,m]|X[c,m]}
\hline
\multirow{2}{*}{Method} & Input & \multicolumn{2}{c|}{Accuracy (IU)} & Time & GPU \\
\cline{3-4}
 & resolution & Category & Class & ($ms$) &  ($MB$)\\
\hline\hline
		& 2048$\times$1024 & 84.0 & 64.0 & 1365 & 5800 \\
\cline{2-6}
FCN-8s	& 1024$\times$512 & 77.9 & 57.4 & 468 & 2220 \\
\cline{2-6}
\cite{longcvpr, longpami} 	& 512$\times$256 & 66.9 & 42.3 & 216 & 1218 \\
\cline{2-6}
		& 256$\times$128 & 50.8 & 23.0 & 162 & 911 \\
\hline 
\end{tabu}
\end{table}

%% file: tbl_analysis_hyperparameter.tex
\begin{table}[!t]
\caption{Analysis on hyperparameters in particle swarm optimization.}
\label{tab:hyperparameters}
\centering
\renewcommand{\arraystretch}{1.1}
\begin{tabu} to 0.48\textwidth { X[c,m]|X[c,m]|X[c,m]|X[c,m]|X[c,m]}
\hline
\multicolumn{2}{c|}{Hyperparameters} & \multirow{2}{*}{\# of nodes} & \multicolumn{2}{c}{Accuracy (IU)} \\
\cline{1-2} \cline{4-5}

Weight & Offset & & Category & Class \\
\hline \hline
\multirow{2}{*}{0.1}		& 0.2 & 1,960 & 43.0 & 17.6  \\
		& 0.3 & 1,984 & 43.1 & 17.6  \\
\hline
\multirow{4}{*}{0.2}		& 0.1 & 1,942 & 42.5 & 16.8  \\
		& 0.2 & 1,954 & 43.2 & 17.6  \\
		& 0.3 & 1,988 & \textbf{43.3} & \textbf{17.8}  \\
		& 0.4 & 1,964 & 43.1 & 17.5  \\
\hline
\multirow{2}{*}{0.3}		& 0.2 & 1,952 & 43.2 & 17.3  \\
		& 0.3 & 1,950 & 43.2 & 17.7  \\
\hline
\end{tabu}
\end{table}

%% file: tbl_analysis_depth.tex
\begin{table}[!t]
\caption{Analysis of the maximum depth in the proposed random forest using the Cityscapes dataset.}
\label{tab:maximumDepthCityscapes}
\centering
\renewcommand{\arraystretch}{1.1}
\begin{tabu} to 0.5\textwidth { c|c|X[c,m]|X[c,m]|X[c,m]|X[c,m]}
\hline
Maximum & \# of  & \multicolumn{2}{c|}{Accuracy (IU)} & Time & GPU \\
\cline{3-4}
depth 	 & nodes & Category & Class & ($ms$) &  ($MB$)\\
\hline\hline
1		& 3 & 16.7 & 5.8 & 1.3 & 72 \\
\hline
5	& 63 & 38.5 & 12.7 & 1.6 & 73 \\
\hline
10 	& 2037 & 47.4 & 19.3 & 2.2 & 74 \\
\hline
15		& 53354 & 51.0 & 22.4 & 3.2 & 91 \\
\hline 
20		& 598317 & 53.9 & 25.3 & 4.6 & 291 \\
\hline 
25		& 2349554 & 56.1 & 27.1 & 6.9 & 938 \\
\hline 
\end{tabu}
\end{table}

%% file: tbl_analysis_category.tex
\begin{table}[!t]
\caption{Analysis of accuracy per category.}
\label{tab:perCategory}
\centering
\renewcommand{\arraystretch}{1.1}
\begin{tabu} to 0.5\textwidth { X[c,m]|X[c,m]|X[c,m]|X[c,m]}
\hline
\multirow{2}{*}{Category} 	 & \multicolumn{3}{c}{Method} \\
\cline{2-4}
 	 		& FCN-8s & ENet & Proposed \\
\hline\hline
Flat			& 98.2 & 97.3 & 93.8 \\
\hline
Nature		& 91.1 & 88.3 & 73.6 \\
\hline
Object 		& 57.0 & 46.8 & 15.7 \\
\hline
Sky			& 93.9 & 90.6 & 84.6 \\
\hline 
Construction	& 89.6 & 85.4 & 61.0 \\
\hline 
Human		& 78.6 & 65.5 & 29.8 \\
\hline 
Vehicle		& 91.3 & 88.9 & 63.3 \\
\hline 
\end{tabu}
\end{table}

%% file: fig_result_cityscapes2.tex
\begin{figure*}[!t] \begin{center}
\begin{minipage}{0.19\linewidth}
\centerline{\includegraphics[scale=0.09]{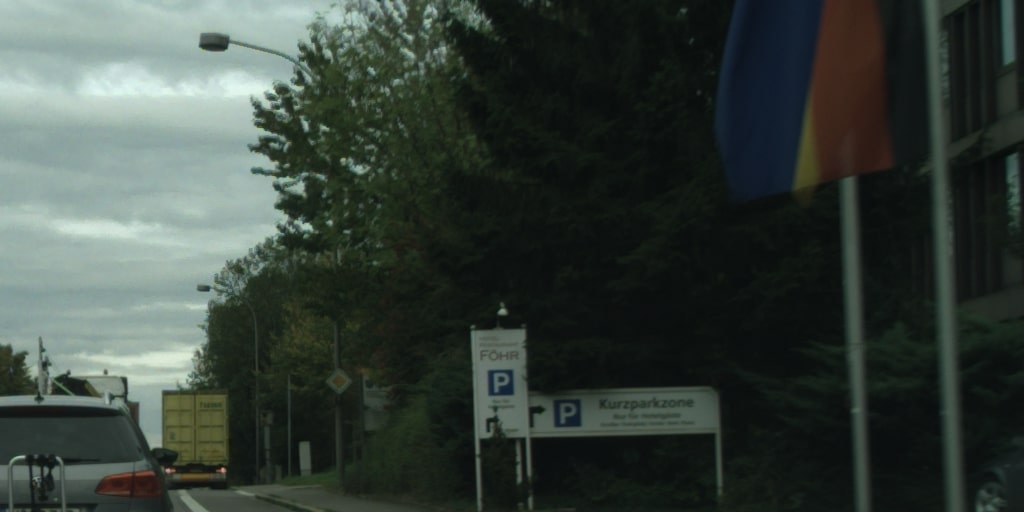}}
\end{minipage}
\begin{minipage}{0.19\linewidth}
\centerline{\includegraphics[scale=0.09]{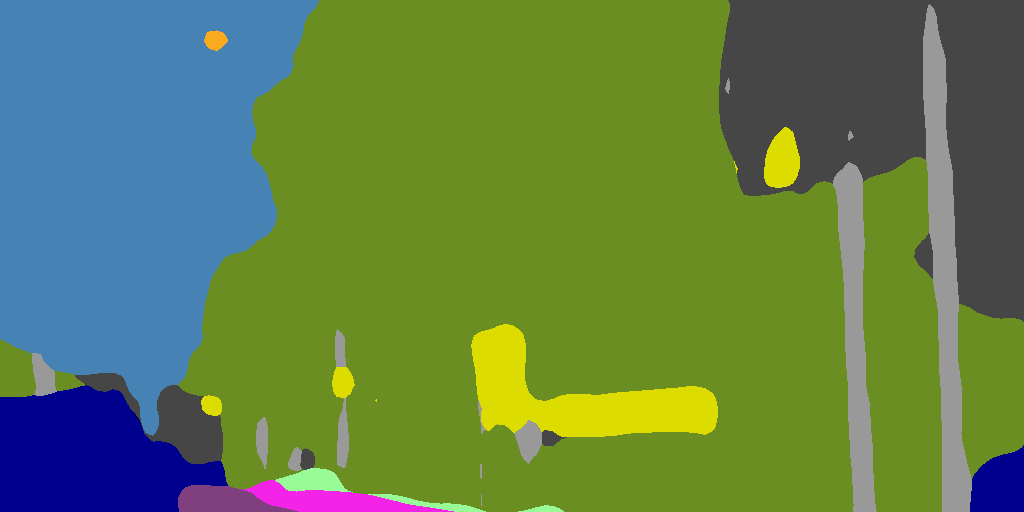}}
\end{minipage}
\begin{minipage}{0.19\linewidth}
\centerline{\includegraphics[scale=0.09]{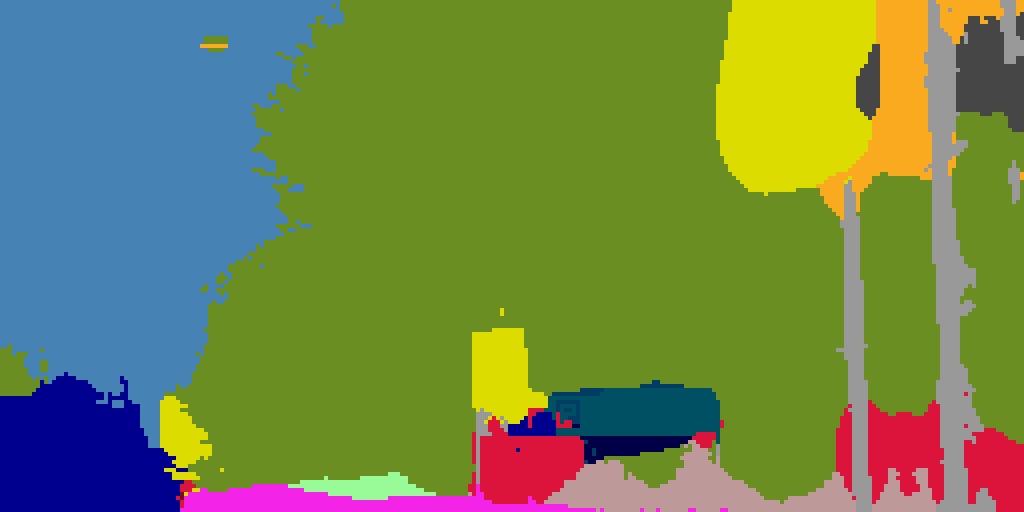}}
\end{minipage}
\begin{minipage}{0.19\linewidth}
\centerline{\includegraphics[scale=0.09]{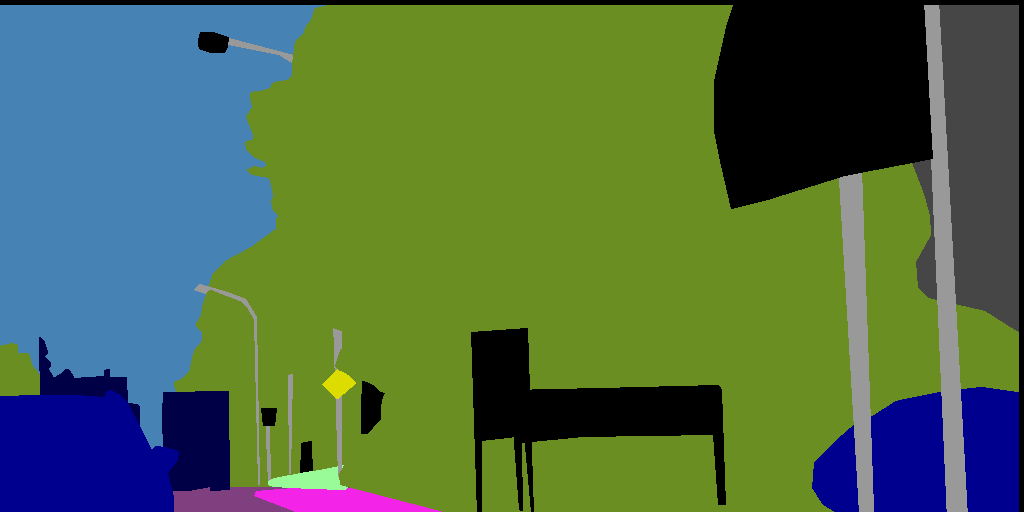}}
\end{minipage}
\\
\vspace{0.1cm}
\begin{minipage}{0.19\linewidth}
\centerline{\includegraphics[scale=0.09]{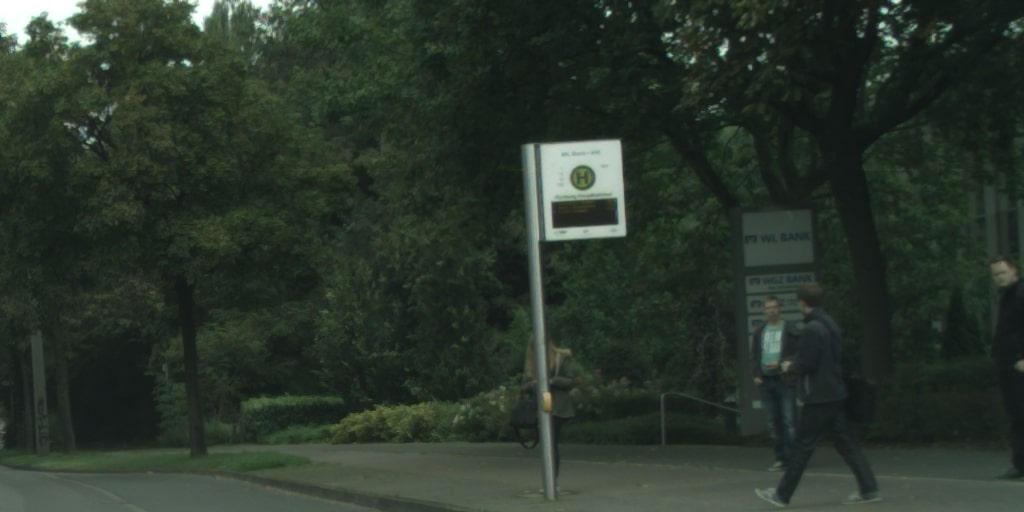}}
\end{minipage}
\begin{minipage}{0.19\linewidth}
\centerline{\includegraphics[scale=0.09]{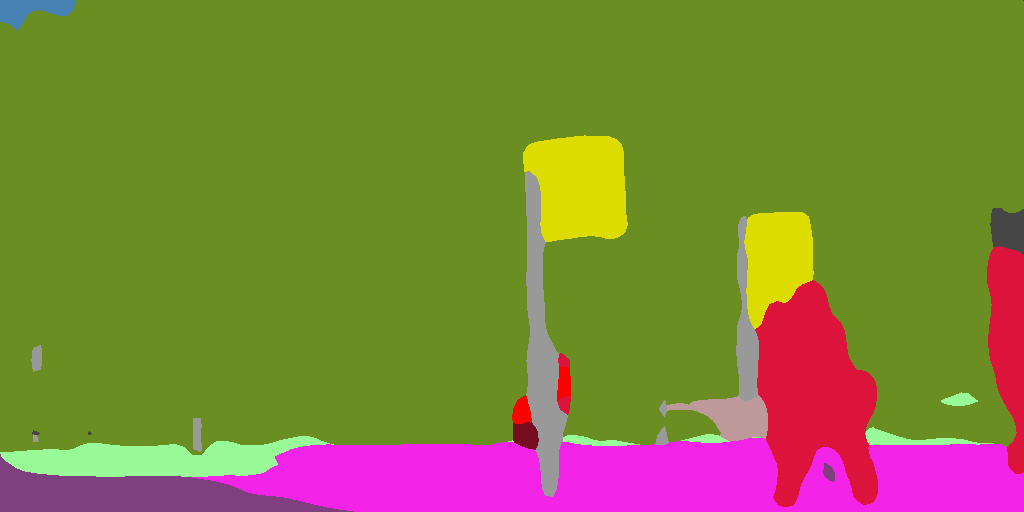}}
\end{minipage}
\begin{minipage}{0.19\linewidth}
\centerline{\includegraphics[scale=0.09]{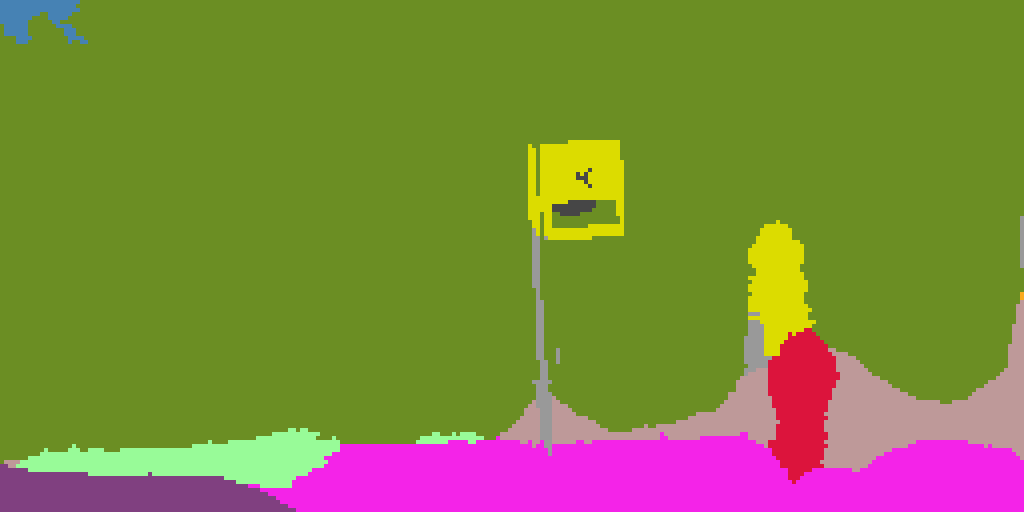}}
\end{minipage}
\begin{minipage}{0.19\linewidth}
\centerline{\includegraphics[scale=0.09]{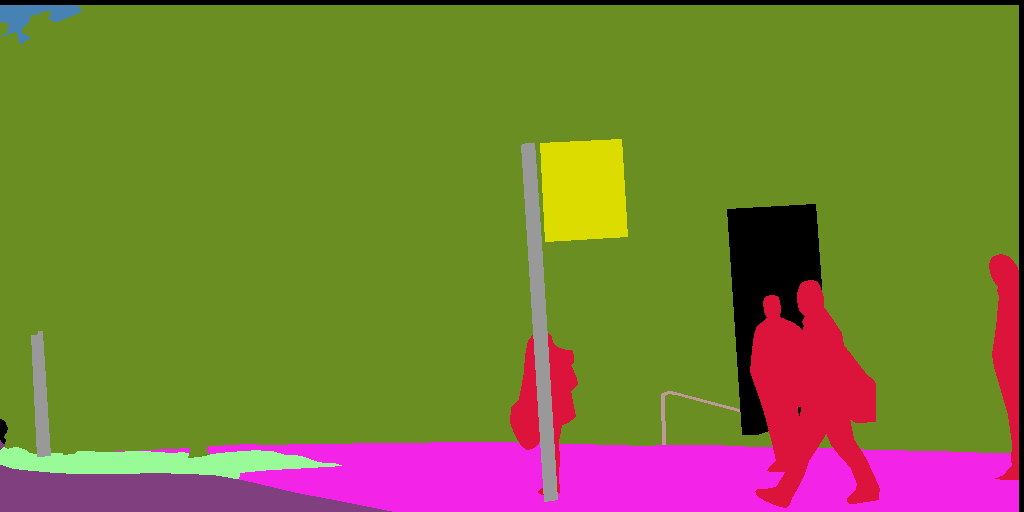}}
\end{minipage}
\\
\vspace{0.1cm}
\begin{minipage}{0.19\linewidth}
\centerline{\includegraphics[scale=0.09]{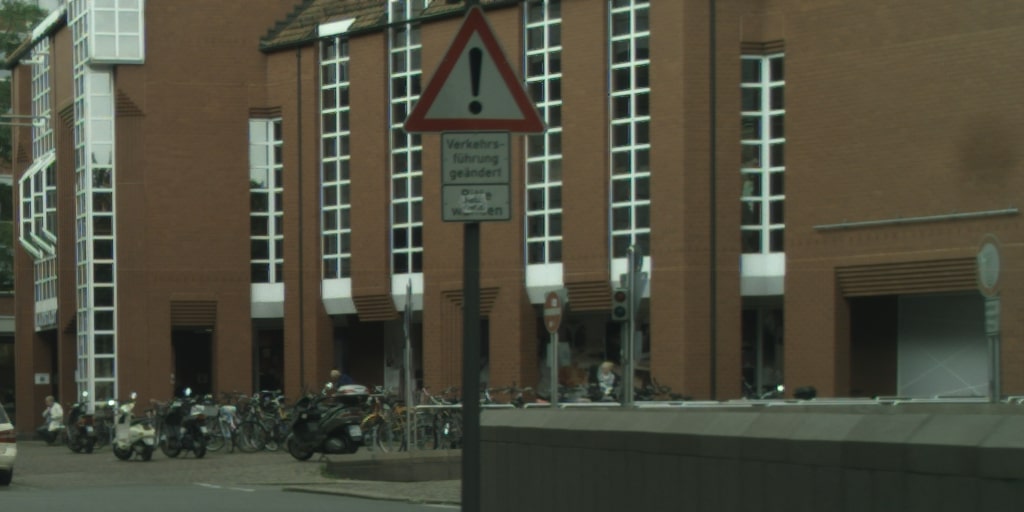}}
\centerline{\footnotesize (a) Input image}
\end{minipage}
\begin{minipage}{0.19\linewidth}
\centerline{\includegraphics[scale=0.09]{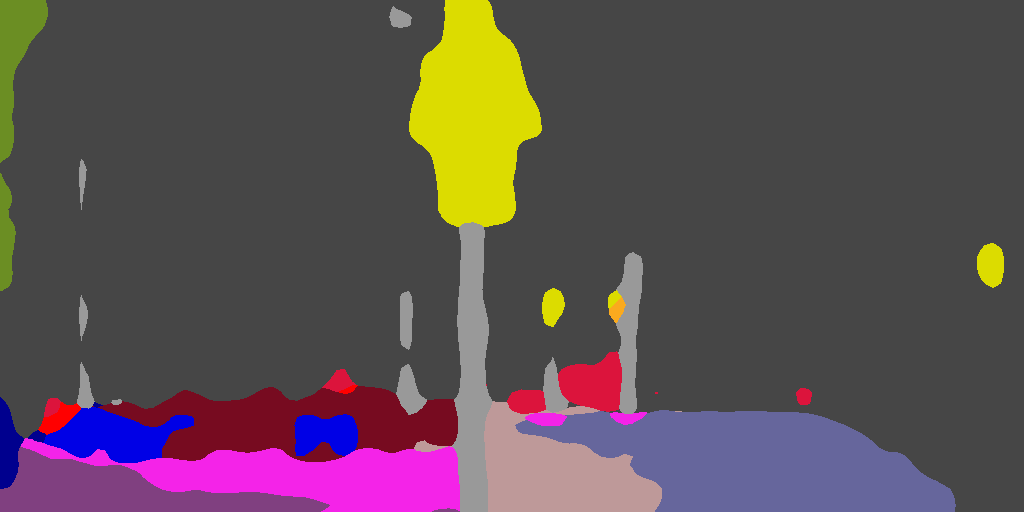}}
\centerline{\footnotesize (b) FCN-8s~\cite{longcvpr, longpami}} 
\end{minipage}
\begin{minipage}{0.19\linewidth}
\centerline{\includegraphics[scale=0.09]{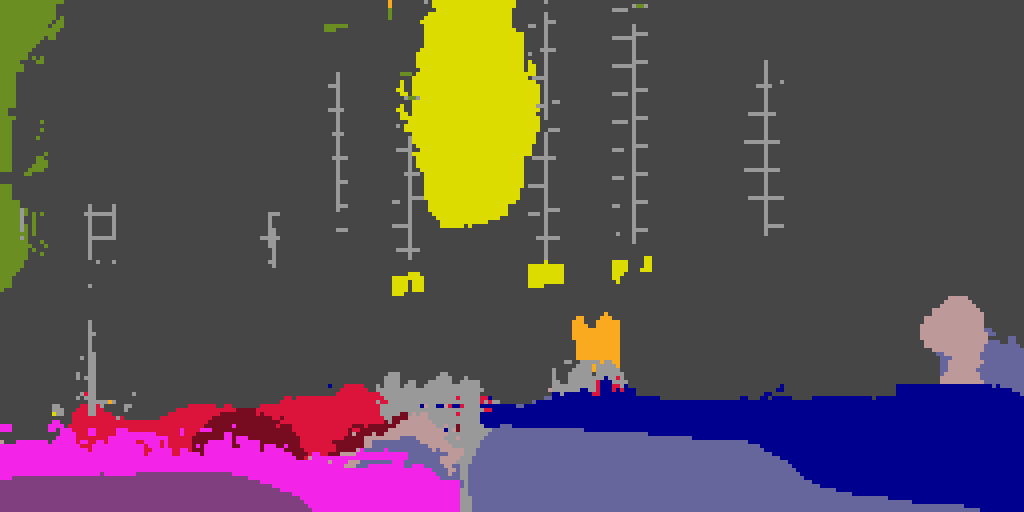}}
\centerline{\footnotesize (c) Proposed method} 
\end{minipage}
\begin{minipage}{0.19\linewidth}
\centerline{\includegraphics[scale=0.09]{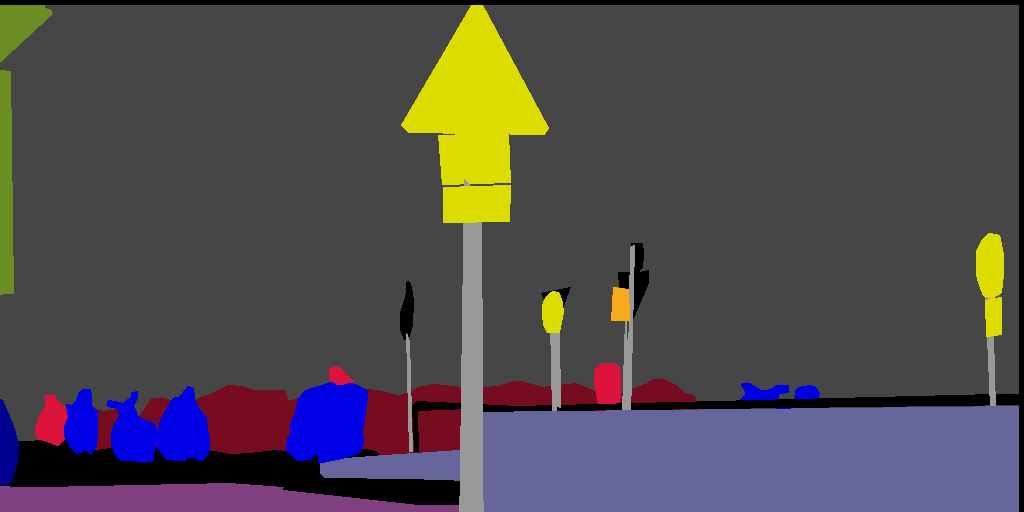}}
\centerline{\footnotesize (d) Ground truth label} 
\end{minipage}
   \caption{The qualitative comparison using enlarged results. Right-top parts of images in~\fref{fig:cityscapesResult} are visualized. (a) Input image. (b) Results of the FCN-8s model~\cite{longcvpr, longpami}. (d) Results of the proposed method using five trees, bilateral filter of 23$\times$23, and the input resolution of 512$\times$256. (e) Ground truth labels.}
\label{fig:cityscapesResult2}
\end{center}\end{figure*}

%% file: tbl_result_nyudv2.tex
\begin{table}[!t]
\caption{The quantitative results of the NYUDv2 dataset.}
\label{tab:nyudv2Result}
\centering
\renewcommand{\arraystretch}{1.1}
\begin{tabu} to 0.5\textwidth { c|X[c,m]|X[c,m]|X[c,m]}
\hline
\multirow{2}{*}{Method} & \multirow{2}{*}{Accuracy} & Time & GPU memory \\
 						&  & ($ms$) & ($MB$) \\
\hline\hline
DeepLab~\cite{Chen16}			& 63.8 & 365 & 1397 \\
Frontend~\cite{YuKoltun2016}		& 62.1 & 457 & 1886 \\
FCN-8s~\cite{longpami}			& 62.1 & 246 & 1573 \\
FCN-16s~\cite{longpami}			& 62.3 & 246 & 1557 \\
FCN-32s~\cite{longpami}			& 61.8 & 251 & 1549 \\
FCN-32s~\cite{longcvpr}			& 60.0 & 251 & 1549 \\
\hline 
Proposed						& 32.6 & 8 & 1080 \\
\hline
\end{tabu}
\end{table}